\documentclass[prodmode,acmtois]{acmsmall}

\usepackage{epstopdf}
\usepackage{algpseudocode}
\usepackage{algorithm}
\usepackage{tikz,pgfplots}
\usepackage{float}
\usepackage{mathtools}
\usepackage{epsfig}
\usepackage{array}
\usepackage{multirow}
\usepackage{afterpage}
\usepackage{amssymb}
\usepackage{ stmaryrd }
\usepackage{csquotes}

\DeclareMathOperator*{\argmax}{arg\,max}

\def\mean#1{\left< #1 \right>}

\acmVolume{0}  
\acmNumber{0}  
\acmArticle{0}  
\acmYear{0}  
\acmMonth{0}

\begin{document}

\markboth{}{Learning Stylometric Representations for Authorship Analysis}

\title{Learning Stylometric Representations for Authorship Analysis}
\author{
STEVEN H. H. DING
\affil{School of Information Studies, McGill University, Canada}
BENJAMIN C. M. FUNG
\affil{School of Information Studies, McGill University, Canada}
FARKHUND IQBAL
\affil{College of Technological Innovation, Zayed University, UAE}
WILLIAM K. CHEUNG
\affil{Department of Computer Science, Hong Kong Baptist University, Hong Kong}
}

\begin{abstract}
Authorship analysis (AA) is the study of unveiling the hidden properties of authors from a body of exponentially exploding textual data. It extracts an author's identity and sociolinguistic characteristics based on the reflected writing styles in the text. It is an essential process for various areas, such as cybercrime investigation, psycholinguistics, political socialization, etc. However, most of the previous techniques critically depend on the manual feature engineering process. Consequently, the choice of feature set has been shown to be scenario- or dataset-dependent. In this paper, to mimic the human sentence composition process using a neural network approach, we propose to incorporate different categories of linguistic features into distributed representation of words in order to learn simultaneously the writing style representations based on unlabeled texts for authorship analysis. In particular, the proposed models allow topical, lexical, syntactical, and character-level feature vectors of each document to be extracted as stylometrics. We evaluate the performance of our approach on the problems of authorship characterization and authorship verification with the Twitter, novel, and essay datasets. The experiments suggest that our proposed text representation outperforms the bag-of-lexical-$n$-grams, Latent Dirichlet Allocation, Latent Semantic Analysis, PVDM, PVDBOW, and word2vec representations.

\end{abstract}






\begin{bottomstuff}
Authors' addresses: Benjamin C. M. Fung (corresponding author), School of Information Studies, McGill University, Montreal, QC, Canada H3A 1X1; email: ben.fung@mcgill.ca
\end{bottomstuff}

\maketitle

\section{Introduction}
\label{sec:introduction}

The prevalence of the computer information system, personal computational devices, and the globalizing Internet have fundamentally transformed our daily lives and reshaped the way we generate and digest information. Countless pieces of textual snippets and documents are generated every millisecond: This is the era of infobesity. Authorship analysis (AA) is one of the critical approaches to turn the burden of a vast amount of data into practical, useful knowledge. By looking into the reflected linguistic trails, AA is a study to unveil an underlying author's identity and sociolinguistic characteristics. The advancement of authorship analysis backed up by stylometric techniques has a fundamental impact on various areas:

\begin{longitem}

\item\emph{Cybercrime investigation}. The distributed nature of cyber space provides an ideal anonymous channel for computer-mediated malicious activities, e.g., phishing scams, spamming, ransom messages, harassment, money laundering, illegal material distribution, etc., because the network-based origins such as IP address can be easily repudiated. Several authorship identification techniques have been developed for the purpose of cyber investigation on SMS text-messaging slips~\cite{DBLP:journals/corr/RagelHS14}, personal e-mails~\cite{DBLP:journals/isci/IqbalBFD13,ding2015visualizable}, and social blogs~\cite{DBLP:conf/sec/YangC14,DBLP:conf/ifip11-9/StolermanOAG14}. Stylometric techniques have been used as evidence in the form of \textit{expert knowledge} in the courts of the UK, the US, and Australia~\cite{juola2006authorship,DBLP:journals/tissec/BrennanAG12}. In a well-known case in the UK, linguistic experts showed that a series of text messages sent from Danielle's phone after she had disappeared were actually not written by her but by her uncle, Stuart Campbell~\cite{granttext}. For more recent information and definitions about linguistic evidence, please refer to the home page of the Institute for Linguistic Evidence (ILE)\footnote{Institute for Linguistic Evidence (ILE): \url{http://linguisticevidence.org/}}.

\item\emph{Marketing, political socialization, and social network analysis}. By inferring the hidden attributes of social network users, businesses and political parties can get a deeper understanding of their audiences and the related discussions. AA techniques have been applied to understand audiences' political preferences~\cite{DBLP:journals/socnet/GolbeckH14,DBLP:conf/icwsm/PennacchiottiP11,golbeck2011computing}, networks of political communication~\cite{DBLP:conf/icwsm/ConoverRFGMF11}, and political learnings of prominent tweeters and media~\cite{DBLP:conf/icwsm/WongTSC13}. Understanding the audiences on social networks helps in planning political strategy~\cite{DBLP:conf/socialcom/ConoverGRFM11} and corrects bias for predicting election results \cite{DBLP:conf/icwsm/WongTSC13}. From the perspective of marketing, it is profitable to identify potential customers by characterizing them and understanding their reviews~\cite{DBLP:journals/jidm/WerenKMMOW14}. \citeN{DBLP:conf/icwsm/PennacchiottiP11} present an AA approach to distinguish whether or not a Twitter user is a fan of Starbucks, which in general can be applied to other types of businesses. Moreover, most of social network analyses are based on some attributes of the user inside the network, and AA techniques can provide more hidden labels for these analyses~\cite{DBLP:conf/icwsm/CohenR13}.

\item\emph{Literary science and education}. Authorship analysis also has a significant impact in the fields related to literary science. Many AA techniques have been developed to infer the disputed authorship of historical documents such as Civil War letters~\cite{klarreich2003bookish}, Shakespeare's plays \cite{klarreich2003bookish}, the Federalist Papers \cite{oakes2004ant,DBLP:conf/eacl/TschuggnallS14,DBLP:conf/coling/NasirGB14}, and the classic French literary mystery: ``Le Roman de Violette"~\cite{DBLP:conf/slsp/BoukhaledG14}. AA techniques are also used to quantify the performance of literary translators, since the best translators will not have their own writing style reflected in the translated works~\cite{DBLP:conf/cosn/AlmishariOT14}. Moreover, AA techniques have been used to understand the personality of students~\cite{DBLP:conf/cosn/AlmishariOT14}, first languages~\cite{DBLP:conf/cosn/AlmishariOT14,DBLP:journals/jasis/TorneyVY12}, and self-reported names~\cite{DBLP:conf/aaaiss/LiuR13}. Furthermore, AA techniques are fundamental to the detection of plagiarism in academic works~\cite{DBLP:journals/corr/CerraDR14,DBLP:conf/irfc/HansenLLA14}.
\end{longitem}

Studies of authorship analysis backed up by computational stylometric techniques can be dated back to the 19th century. Many customized approaches focusing on different sub-problems and scenarios have been proposed~\cite{stamatatos2009survey}. It has been a successful line of research~\cite{DBLP:journals/tissec/BrennanAG12}. Research problems in authorship analysis can be broadly categorized into three types: \emph{authorship identification} (i.e., identify the most plausible author given a set of candidates~\cite{DBLP:journals/isci/IqbalBFD13,ding2015visualizable}), \emph{authorship verification} (i.e., verify whether or not a given candidate is the actual author of the given text~\cite{DBLP:conf/clef/StamatatosDVSPJSB14}), and \emph{authorship characterization} (i.e., infer the sociolinguistic characteristics of the author of the given text~\cite{rangel2014overview}). Both the problems of authorship identification and authorship characterization can be formulated as a one-class text classification problem. For the authorship attribution problem, the classification label is the identity of the anonymous text snippet; for the authorship attribution problem, the label can be the hidden properties of the anonymous author, such as age and gender. There exists other variances of the authorship attribution problems, such as the open-set problem, the closed-set problem, and the attribution-with-probability-output problem~\cite{stamatatos2009survey,DBLP:journals/isci/IqbalBFD13,ding2015visualizable}.

\begin{figure*}
\centering
\includegraphics[width=5in]{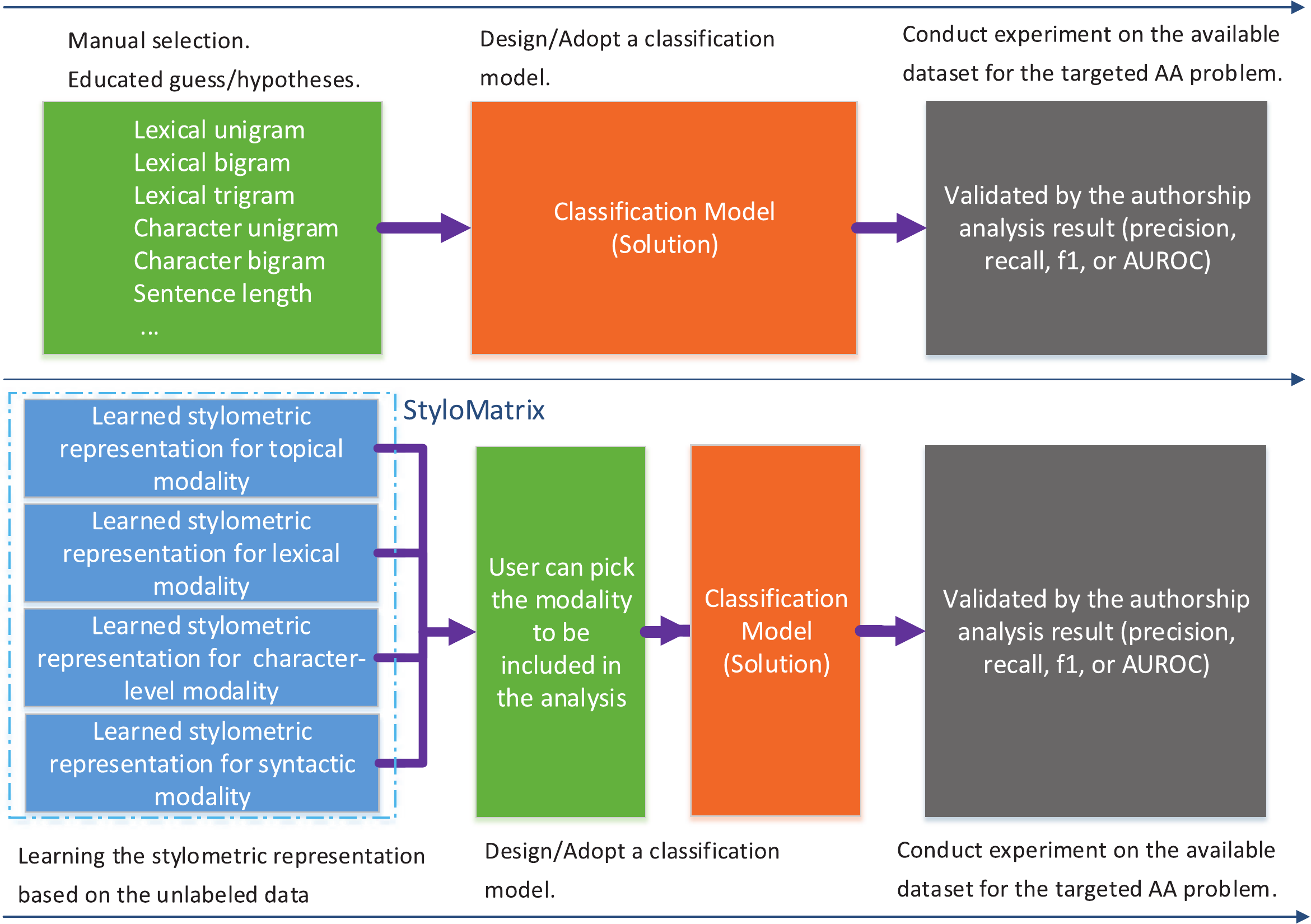}
 
\caption{Overview of the traditional solution and the proposed solution for authorship analysis.}
\label{fig:problemFlow}
\end{figure*}

Regardless of the studied authorship problems, the existing solutions in previous AA studies typically consist of three major processes, as shown in the upper flowchart of Figure~\ref{fig:problemFlow}): the feature engineering process, the solution design process, and the experimental evaluation process. In the first, a set of features are manually chosen by the researchers to represent each unit of textual data as a numeric vector. In the second process, a classification model is carefully adopted or designed. At the end, the entire solution is evaluated based on the specific datasets. Representative solutions are~\citeN{DBLP:conf/emnlp/BurgerHKZ11}, \citeN{DBLP:journals/corr/NirkhiD14}, and~\citeN{DBLP:conf/ciarp/CavalcanteRC14}. Exceptions are few recent applications of the topic models that actually combine these two process into one~\cite{DBLP:conf/icdm/PratanwanichL14,DBLP:journals/ipm/Savoy13,DBLP:journals/coling/SeroussiZB14}. Still, the two-processes-based studies on authorship analysis problems dominate~\cite{rangel2014overview}.

During the feature engineering process, given the available dataset and application scenario, authorship analysts manually select a broad set of features based on the hypotheses or educated guesses, and then refine the selection based on the experimental feedback. As demonstrated by previous research~\cite{DBLP:journals/tois/Savoy12,DBLP:conf/clef/ZamaniEBADS14,DBLP:conf/sac/Savoy13,ding2015visualizable}, the choice of the feature set (i.e., the feature selection method) is a crucial indicator of the prediction result, and it requires explicit knowledge in computational linguistics and tacit experiences in analyzing the textual data. Manual feature engineering is a time-consuming and labor-intensive task. The hand-crafted feature representations have very limited generalizability on different data and scenarios. We have shown that the amount of data and the complexity of the problem to be solved have a strong implication on the analysis result~\cite{ding2015visualizable} even when a full set of $n$-grams is employed. Thus, the generalizability and sensitivity to different application scenarios is critical.

Manual feature engineering to a high degree limits the potentials of the whole available feature space because it makes a strong simplification on the language model. For example, the \emph{bag of words} (\emph{BoW}) model, in which the text is represented as an unordered list of words with their respective frequency value, ignores the dependency between words and the sequential information. The lexical $n$-gram model describes a feature as a consecutive $n$ lexical token; it can capture the co-occurrence relationship within an $n$-length neighbor. Still, it is unable to capture the contextual relationship between words over long sentences. Also, most feature selection methods trim the original space based on a specific metric, for example, by picking the top-$k$ frequent lexical $n$-gram. In our previous study~\cite{ding2015visualizable}, we showed that using a full set of $n$-gram features greatly promotes the accuracy of authorship identification, which indicates the reduced classification power of feature selection for authorship analysis.

Inspired by the recent development of the unsupervised representation learning in deep learning~\cite{DBLP:journals/corr/abs-1301-3781}, we raise two new research questions for authorship analysis. (1) Given the \emph{unlabeled} textual data, can we automatically come up with a vectorized numeric representation of the writing style? (2) Can we contribute to the AA field by discovering and generalizing new, special, and interpretable linguistic features from the available data instead of validating the educated guess and hypotheses on the data?

In this paper, we present a stylometric representation learning approach for authorship analysis (AA). Refer to the lower flowchart in Figure~\ref{fig:problemFlow}. The goal is to learn an effective vector representation of writing styles of different linguistic modalities to mitigate the aforementioned issue in AA study. Following the previous work~\cite{1:16:modality,sapkota2013use,ding2015visualizable}, we use the concept \emph{linguistic modalities} to denote the categories of linguistic features \cite{1:16:modality}. We broadly categorize them into four modalities: the \emph{topical modality}, the \emph{lexical modality}, the \emph{character-level modality}, and the \emph{syntactic modality}. It is noted that the term ``modality'' used here is different from the term ``multi-modality'' in machine learning. The first one denotes a category of linguistic features, and the latter denotes a combination of different ways in which information is presented, such as text, image, rating, etc. In this paper, ``modality'' and ``linguistic modality'' are used interchangeably to denote the categories of linguistic features. Also, we use the term \emph{representation} and \emph{embedding} interchangeably to describe the vectorized representation of feature. In the first stage, we learn the stylometric representation for different linguistic modalities based on the unlabeled textual data. At the second stage, an authorship analyst can select the modality according to his or her needs. If the scenario requires the least interference from the topic-related information, the analyst can discard the topical modality, or, more strictly, both the topical and lexical modality. Such a design inherits the flexibility of the original hand-crafted stylometric features while it enables the learned representation to fit into the available data.

The basic idea of our proposed solution is to simulate how people construct a sentence based on different linguistic modalities. The proposed approach follows the recent ideas and models of estimating word embedding~\cite{DBLP:journals/corr/abs-1301-3781} and paragraph embedding~\cite{DBLP:journals/corr/LeM14} to efficiently approximate the factorization of different co-occurrence matrices. The proposed approach can be applied to any authorship analysis studies that involve feature engineering processes.

To the best of our knowledge, this is the very first work attempting to automate the feature engineering process and discover the stylometric representations for authorship analysis. Specifically, our major contributions are summarized as follows:

\begin{itemize}
\item We propose a different solution flow for authorship analysis. Instead of manually engineering the stylometric features, we learn the representation of the writing style based on the available unlabeled texts according to different linguistic modalities. The user/researcher can pick the modalities based on their needs and interests in the context of the authorship analysis problem. For example, political socialization researchers are interested in content, so they may choose topic-modality. In contrast, cybercrime investigators would prefer avoiding topic-related features since given a harassment letter, the candidate authors may not have previously written anything on such a topic.

\item We propose a joint learning model that can learn simultaneously the distributed word representation as well as the topical bias and lexical bias representations of each document based on unlabeled texts. The joint learning model simulates the sentence composition process and captures the joint effects of topical bias and lexical bias in picking a specific word to form the sentence. The learned topical vector representation of a document is able to capture the global topical context, while the learned lexical representation of a document is able to capture the personal bias in choosing words under the given global context.

\item Using a similar learning representation approach, we propose how the character-level and syntactic-level representations of each document can be learned. The character-level model captures the morphological and phonemes bias of an author when he/she is composing a lexical token. It models the probability of how a character element is chosen to form a given lexical token in the sentence. The syntactic-level model captures the syntactic/grammatical bias of an author when he/she is putting words together to construct a sentence. It models a prediction path that maximally avoids the dependency information introduced by the Part-of-Speech (POS) tagger. To the best of our knowledge, this paper proposes the first model that learns a character-level representation and the first model that learns a syntactic-level representation.

\item
We evaluate the effectiveness of the learned representations as stylometrics via extensive experiments and show its superiority over the state-of-the-art representations and algorithms for author verification and author characterization tasks. Without using any labels we achieve the best result on the PAN 2014~\cite{DBLP:conf/clef/StamatatosDVSPJSB14} English authorship verification problem with respect to the Area Under Receiver Operating Characteristic curve (AUROC). By using a simple logistic regression classifier over the learned representations, we achieve the best result on the ICWSM 2012~\cite{al2012homophily} authorship characterization dataset without any social network related structural information. The characteristics include age range, gender, and political orientation.
\end{itemize}

The rest of this paper is organized as follows: Section~\ref{sec:mining} elaborates our stylometric learning models and discusses their relation to the existing stylometric features according to different linguistic modalities. Section~\ref{sec:verification} elaborates our evaluation of the proposed models on the authorship verification problem with the PAN 2014 dataset. Section~\ref{sec:characterization} presents our evaluation of the proposed models on the problem of authorship characterization with the ICWSM 2012 dataset. Relevant works are situated throughout the discussions in this paper. Finally Section~\ref{sec:conclusion} concludes this paper and explores future directions.

\section{Mining Stylometric Representations for Authorship Analysis}
\label{sec:mining}
In this section, we present the proposed models for learning the stylometric representations on unlabeled training data and estimating the representations for unseen testing data. To start with, we will define several key concepts and the studied stylometric learning problem.

To be consistent in terminology, \emph{text dataset} refers to the union of available labeled and unlabeled text; \emph{document} and \emph{writing sample} are used interchangeably to refer to the minimum unit of text data to be analyzed. A writing sample consists of a list of sentences, and a sentence consists of a sequence of lexical tokens. Each lexical token has its respective POS tag in the corresponding sentence.

This section corresponds to the first process of the lower flowchart in Figure~\ref{fig:problemFlow}, where only unlabeled text data are available. In this process we learn the representation of each chosen unit of text into four vectorized numeric representations, respectively, for four linguistic modalities. We formally define the stylometric feature learning problem as follows:

\begin{definition}{(\textit{stylometric representation learning})}\label{def:pro:stylometricRepresentationLearning}
The given text dataset is denoted by $\mathbb{D}$, and each document is formulated as $\omega \in \mathbb{D}$. A document $\omega$ consists of a list of ordered sentences $\cal{S}(\omega) = s[1:a]$, where $s_a$ represents one of them. Each sentence consists of an ordered list of lexical tokens $\cal{T}(s_a) = t[1:b]$, where $t_b$ represents the token at index $b$. $\cal{P}(t_b)$ denotes the Part-of-Speech tag for token $t_b$. Given $\mathbb{D}$, the task is to learn four vector representations $\vec{\theta}^{tp}_{\omega} \in \mathbb{R}^{\cal{D}(tp)}$, $\vec{\theta}^{lx}_{\omega} \in \mathbb{R}^{\cal{D}(lx)}$, $\vec{\theta}^{ch}_{\omega} \in \mathbb{R}^{\cal{D}(ch)}$, and $\vec{\theta}^{sy}_{\omega} \in \mathbb{R}^{\cal{D}(sy)}$, respectively, for topical modality $tp$, lexical modality $lx$, character-level modality $ch$, and syntactic modality $sy$ for each document $\omega \in \mathbb{D}$. $\cal{D}(\cdot)$ denotes the dimensionality for a given modality.~$\blacksquare$
\end{definition}

We argue that the division of the whole feature space according to linguistic modalities is necessary because different application scenarios have different requirements of the features. Moreover, even for manual feature engineering, writing styles that correspond to different linguistic dimensions are constructed differently by humans \cite{1:16:modality,DBLP:journals/jasis/TorneyVY12}; therefore, they need to be grouped together.

Take the typical dynamic bag-of-lexical-$n$-gram model, with ranking by frequency, as a lexical modality representation example. Given a text dataset $\omega \in \mathbb{D}$, the top $k$ lexical $n$-grams $\cal{G}(\omega) = g[1:c]$ are selected based on their occurring frequency. $g_c$ represents one of them. For each document $\omega$, a lexical level modality representation is constructed as $\vec{\theta}^{k}_{\omega} \in \mathbb{R}^{k}$, where $\vec{\theta}^{k}_{\omega}[c]$ is the frequency value of $g_c$ in $\omega$. In the following section, we describe the proposed models for stylometric learning according to different linguistic modalities.

\subsection{Joint learning model for topical modality and lexical modality}
\label{sec:topic}

In this section we are interested in both the topical modality and the lexical modality; both operate on the lexical tokens of a document. To start with, we look into their nature and the existing AA solutions that involve these two modalities.

\subsubsection{Topical modality}

The topical modality concerns the differences of interested topics reflected from the plain text. For example, in web blogs males may talk more about the information technologies, while females may talk more about fashion and cosmetics. It is widely acknowledged that the topics reflected from text actually depend on the distribution of words or combination of words~\cite{6:2:topicwordrelationship}.

Typical LDA-based AA techniques~\cite{DBLP:journals/ipm/Savoy13,DBLP:conf/icdm/PratanwanichL14,DBLP:journals/coling/SeroussiZB14} construct the topics by drawing a distribution over the lexical tokens and then represent the writing style as the distribution over topics. For this modality it is intuitive to directly use the LDA or the LSA topic models and represent the constructed distribution over topics as the stylometric representation. However, it has been shown that the representation of words learned by a neural network performs significantly better than the LSA model, and the LDA model becomes computationally infeasible on large datasets~\cite{DBLP:journals/corr/abs-1301-3781}. Thus, we seek to customize existing embedding learning neural network models to combine the co-occurrence relationship between words and document and the co-occurrence relationship between words.

\subsubsection{Lexical modality}

The lexical modality is concerned with the specific choice of words, given the context. Different from the topical modality, in which the stylometric representation captures the relationship between word and document, the lexical modality captures the difference in the co-occurrence relationship between words reflected by the text.

The most widely employed lexical features are function words. They have been shown to be effective for capturing the differences in writing styles~\cite{DBLP:journals/lalc/KoppelAS02}. Usually the frequency value of the function words are used to represent the features~\cite{DBLP:conf/kes/Baron14,pan:halvanivebav,DBLP:journals/cas/HaCohen-KernerM14}. They are effective for identifying the first language of the authors~\cite{DBLP:journals/jasis/TorneyVY12,DBLP:journals/cacm/ArgamonKPS09}, identifying the actual author of French literature~\cite{DBLP:conf/slsp/BoukhaledG14}, and characterizing the gender of e-mails~\cite{DBLP:conf/acsac/CorneyVAM02}. Especially for the first language detection, the effect of language transfer affects the use of function words in the secondary language~\cite{DBLP:journals/jasis/TorneyVY12}. These feature representations can be regarded as the typical ``bag-of-words'' model. \citeN{DBLP:journals/corr/SegarraER14} present the function words differently. They construct a function word adjacent matrix to capture the relationships among function words, and model the probability by regarding it as a Markov chain. Their approach considers the co-occurring relationship among function words.

Another useful type of lexical features are lexical $n$-grams, which denote a sequence of consecutive words of length $n$. Lexical $n$-grams are becoming popular as they are shown to be more effective than character $n$-grams and syntactic $n$-grams when all the possible $n$-grams are used as features~\cite{ding2015visualizable}. Moreover, it has been shown to be effective in identifying the gender of tweeters~\cite{DBLP:conf/emnlp/BurgerHKZ11}. However, the study presented by~\citeN{rao2010detecting} shows that the socio-linguistic features outperform the lexical $n$-gram approach ($n \in {1,2}$) when characterizing gender and age, but when characterizing the region of the Twitter user, the $n$-gram-based approach out-performs the socio-linguistic feature. This is possibly because people in different regions discuss different topics.

The lexical $n$-gram approach has two problems. First, the current $bag-of-ngram$ features fail to capture the co-occurring relationship between words in a longer context due to the limit of the parameter $n$ and the independence assumption of the $n$-grams. Simple unigram (i.e., $n=1$) and bigram (i.e., $n=2$) features can hardly capture the relationship among nouns across the whole sentence, and the relationship between each bigram/trigram is considered independent. Second, the current $n$-gram approach heavily depends on the feature selection method~\cite{DBLP:conf/clef/ZamaniEBADS14,DBLP:journals/ipm/Savoy13,DBLP:journals/jql/Pavlyshenko14}. The space of the complete $n$-gram ($n \in \mathbb{N}$) features is indeed sparse and can be greatly compressed for the problem of authorship analysis. Most of the AA studies simply discard some $n$-gram features based on the threshold of a specific measure, such as occurring frequency; a better way is to analyze the principal component or conduct the factorization of the co-occurrence matrix. However, it is computationally infeasible to do a full factorization for a large dataset.

\subsubsection{Joint modeling of topical and lexical modalities} Both the topical modality and the lexical modality operate on the lexical tokens. A text document $\omega$ can be considered to be generated by the author under a mixture effect of topical bias and lexical bias. In the bag-of-$n$-grams approaches for AA studies, it is difficult to distinguish whether an $n$-gram selected to form a sentence is mainly due to the holistic topics of the document or the personal lexical preference. The LDA-based and LSA-based approaches fail to consider the lexical preference; they only considered the co-occurrence relationship between documents and words instead of the co-occurrence relationship between words.

In order to best separate the mixed effects of topical modality and lexical modality, and to address the aforementioned issues, we propose a joint learning model in which a document is considered as a lexical token picking process and the author picks tokens from his/her vocabulary in sequence to construct sentences and express what his/her interests are. We consider three factors in this token picking process: the topical bias, the local contextual bias, and the lexical bias.

\begin{itemize}
\item \emph{Topical bias.} Based on the certain holistic meaning (i.e., topics) to be conveyed through the text, the author is limited to a vague set of possible thematic tokens. For example, if the previously picked tokens are mostly about Microsoft, then the author will have a higher chance of picking the word ``Windows'' in the rest of the document because they are probably under a similar topic. Given the topics of the document, the author's selection of the next token in a sentence is influenced by the related vocabulary under these topics.

\item \emph{Local contextual bias.} A document has both holistic topics and local contexts. Both influence how the next word is chosen in a sentence. For example, a document about Microsoft may consist of several parts that cover its different software products. Moreover, the context can be irrelevant to the topic. For example, a web blog may have an opening about weather that has nothing to do with the holistic topic in the following text.

\item \emph{Lexical bias.} Given the topics and their related vocabularies, the author has different choices for picking the next token to convey a similar meaning. For example, if the author wants to talk about the good weather, He or she may pick the adjective ``nice'' to describe the word ``day''. Alternatively, the author can pick other words such as ``great'', ``wonderful'', ``fantastic'', or ``fabulous'', etc. The variations in choosing different words to convey a similar meaning introduce the lexical bias for an author to construct the document.
\end{itemize}

The word picking process is a sequence of individual decision problems influenced by the individual topical bias, contextual bias, and lexical bias; therefore, it is natural to jointly learn the topical representation and lexical representation in the same model. It has the advantage of modeling their joint effects simultaneously and at best of minimizing the interference between the learned representations.

\subsubsection{The proposed joint model} This section introduces our proposed joint learning model for the topical modality and lexical modality. The goal is to estimate $\vec{\theta}^{tp}_{\omega} \in \mathbb{R}^{\cal{D}(tp)}$ and $\vec{\theta}^{lx}_{\omega} \in \mathbb{R}^{\cal{D}(lx)}$ in Definition~\ref{def:pro:stylometricRepresentationLearning}.

\begin{figure*}
\centering
\includegraphics[width=5in]{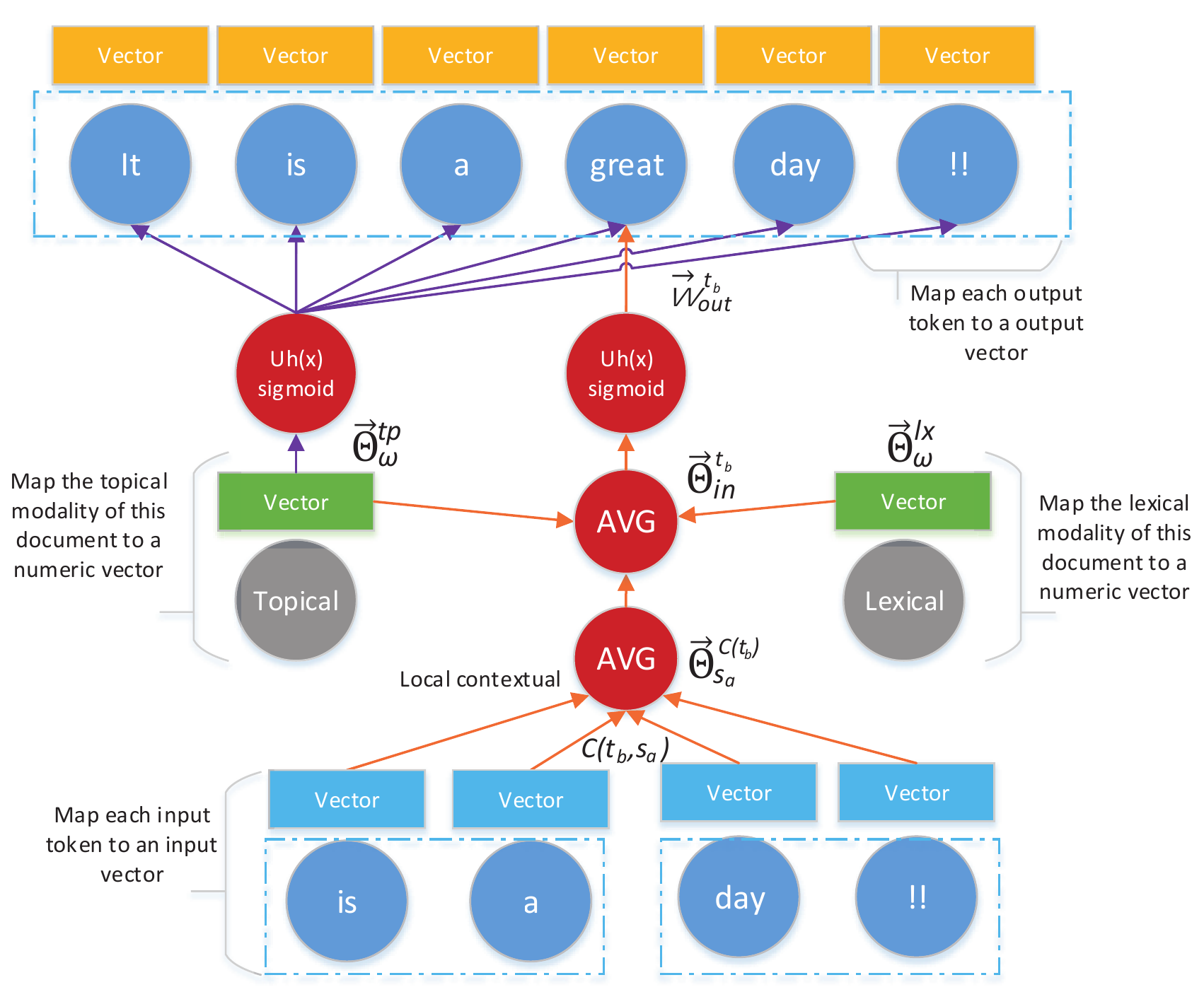}
\caption{The joint model for learning the stylometric representation of the topical and lexical modalities.}
\label{fig:topic}
\end{figure*}

Figure~\ref{fig:topic} depicts the model, which is a single-layer neural network with two output layers. The input is the joint effect of topical bias, local contextual bias, and lexical bias. Recall that the contextual bias concerns the local information surrounding the token to be picked. We represent the vectorized local contextual bias surrounding token $t_b$ in its corresponding sentence $s_a$ as $\theta^{\cal{C}(t_b)}_{s_a}$. The first output is the prediction probability of the targeted word to be chosen by the author. The model tries to maximumize the log probability for the first output:

\begin{align}
\label{equ:lexical:j1}
\argmax J_1(\theta) = \argmax \frac{1}{|\mathbb{D}|}\sum^{\mathbb{D}}_{\omega}\sum^{\cal{S}(\omega)}_{s_a}\sum^{\cal{T}(s_a)}_{t_b} log \; \mathbf{P}(t_b| \underbrace{\vec{\theta}^{tp}_{\omega}}_\text{topical}, \underbrace{\vec{\theta}^{lx}_{\omega}}_\text{lexical}, \underbrace{\theta^{\cal{C}(t_b)}_{s_a}}_\text{contextual})
\end{align}

Similar to the other neural-network-based paragraph/word embedding learning models \cite{def:softmax:h,DBLP:journals/corr/abs-1301-3781,DBLP:journals/corr/LeM14}, this model maps each lexical token $t_b$ into two vectors: $\vec{w}^{t_b}_{in} \in \mathbb{R}^{dw}$ (the yellow rectangles in Figure~\ref{fig:topic}) and $\vec{w}^{t_b}_{out} \in \mathbb{R}^{dw}$ (the blue rectangles in Figure~\ref{fig:topic}) where $dw$ denotes the dimensionality. $\vec{w}^{t_b}_{in}$ is used to construct the input of contextual bias for the neural network, and $\vec{w}^{t_b}_{out}$ is used for the multi-class prediction output of the neural network. They are all model parameters to be estimated on the text data.

The local context of a token is represented by its surrounding tokens within the window size. Given a token $t_b$ in a sentence $s_a$ with a sliding window of size $\cal{W}(tp)$, the context of $t_b$ is formulated as $\cal{C}(t_b, s_a) = \{t_{b-\cal{W}(tp)}, \ldots, t_{b-1}, t_b, t_{b+1}, \ldots, t_{b+\cal{W}(tp)}\}$ where $\cal{C}(t_b, s_a) \subseteq \cal{T}(s_a)$. The contextual bias input to the neural network is defined as the average over the input mapped vectors of $\cal{C}(t_b)$. We define $\mean{\cdot}$ as the vector element-wise average function:

\begin{equation}
\label{equ:lexical:context}
\theta^{\cal{C}(t_b)}_{s_a} = \mean{\sum^{\cal{C}(t_b, s_a)}_{t}\vec{w}^{t}_{in}}
\end{equation}

The other two inputs to the model are the topical bias $\vec{\theta}^{tp}_{\omega} \in \mathbb{R}^{\cal{D}(tp)}$ and the lexical bias $\vec{\theta}^{lx}_{\omega} \in \mathbb{R}^{\cal{D}(lx)}$. In order to have the model working properly, we need to set $\cal{D}(lx)$, $\cal{D}(tp)$, and $dw$ equal to $d_1$, where $d_1$ is the parameter of the whole model that indicates the dimensionality for both the lexical modality representation and topical modality representation. With these three input vectors we further take their average as joint input vector $\theta^{t_b}_{in}$ since it is costly to have a fully connected layer to have the model trained in a reasonable time.

\begin{equation}
\label{equ:lexical:in}
\vec{\theta}^{t_b}_{in} = \mean{\underbrace{\vec{\theta}^{tp}_{\omega}}_\text{topical} + \underbrace{\vec{\theta}^{lx}_{\omega}}_\text{lexical} + \underbrace{\vec{\theta}^{\cal{C}(t_b)}_{s_a}}_\text{contextual}}
\end{equation}

\begin{example}
\label{example:topical-lexical}
Consider a simple sentence: $t_a = $ ``it is a great day !!'' in Figure~\ref{fig:topic}. For each token $\{t_b|b \in [1,6]\}$ we pass forward the neural network. We take $b=4$ and $t_b=$'great' for example. The training process is the same for other values of $b$. Given a windows size of $2$, we construct the local context of $t_4$ as $\cal{C}(t_4, s_a) = \{t_2,t_3,t_5,t_6\} = \{$`is', `a', `day', `!!'$\}$. We map these two tokens into their corresponding vector representations $\vec{w}^{t_2}_{in}$, $\vec{w}^{t_3}_{in}$, $\vec{w}^{t_5}_{in}$ and $\vec{w}^{t_6}_{in}$. With $\vec{\theta}^{tp}_{\omega}$ and $\vec{\theta}^{lx}_{\omega}$, we calculate $\vec{\theta}^{t_4}_{in}$ using Equation~\ref{equ:lexical:in}.  $\blacksquare$
\end{example}

Suppose that we use the typical soft-max multi-class output layer. The first output of this model captures the probability of picking a word $t_b$ based on the joint bias input $\theta^{t_b}_{in}$ as follows:
\begin{equation}
\begin{split}
\label{equ:lexical:p1}
\mathbf{P}(t_b| \underbrace{\vec{\theta}^{tp}_{\omega}}_\text{topical}, \underbrace{\vec{\theta}^{lx}_{\omega}}_\text{lexical}, \underbrace{\vec{\theta}^{\cal{C}(t_b)}_{s_a}}_\text{contextual}) &=  \mathbf{P}(\vec{w}^{t_b}_{out}|\vec{\theta}^{t_b}_{in}) =  \frac{f(\vec{w}^{t_b}_{out}, \vec{\theta}^{t_b}_{in})}{\sum_{t}^{V}f(\vec{w}^{t}_{out}, \vec{\theta}^{t_b}_{in})} \\
f(\vec{w}^{t}_{out},\vec{\theta}^{t_b}_{in}) &= Uh((\vec{w}^{t}_{out})^T \times \vec{\theta}^{t_b}_{in})
\end{split}
\end{equation}	
$V$ denotes the whole vocabulary constructed upon the text dataset $\mathbb{D}$. $Uh(\cdot)$ denotes the element-wise sigmoid function. It corresponds to the red circle under the first output in the Figure~\ref{fig:topic}. This function scales the output to the range of $[0, 1]$, so its output can be interpreted as probability. $\vec{w}^{t}_{out}$ is the mapped out vector for the lexical token $t$.

By substituting the log probability in Equation~\ref{equ:lexical:j1} with Equation~\ref{equ:lexical:p1} and taking derivatives respectively on $\vec{w}^{t}_{out}$ and $\vec{\theta}^{t_b}_{in}$, we have the gradients to be updated for each $t_b$ at each mini-batch in the back propagation algorithm that is used to train this model:

\begin{equation}
\label{equ:lexical:p1:derivative}
\begin{split}
\frac{\partial }{\partial \vec{w}^{t}_{out}} J(\theta)_1 &= \Big(\llbracket t == t_b\rrbracket - \mathbf{P}(\vec{w}^{t}_{out}|\vec{\theta}^{t_b}_{in})\Big) \times \vec{\theta}^{t_b}_{in} \\
\frac{\partial }{\partial\theta^{t_b}_{in}} J(\theta)_1 &= \vec{w}^{t_b}_{out} - \sum^{V}_{t} \mathbf{P}(\vec{w}^{t}_{out}|\vec{\theta}^{t_b}_{in}) \times \vec{w}^{t}_{out}
\end{split}
\end{equation}	
$\llbracket \cdot \rrbracket$ is an identity function. If the expression inside this function is evaluated to be true, then it outputs 1; otherwise 0. For example, $\llbracket1 + 2 == 3\rrbracket = 1$ and $\llbracket1 + 1 == 3\rrbracket = 0$.

However, using a full soft-max layer is costly and inefficient because for each $t_b$ we need to update $d_1 \times (|V| + 2 + \cal{W}(tp) \times 2)$ parameters, where $|V|$ can be large. Following recent development of an efficient word embedding learning approach~\cite{DBLP:journals/corr/abs-1301-3781}, we use the negative sampling method to approximate the complete soft-max layer:
\begin{equation}
\begin{split}
\label{equ:lexical:p2}
log \; \mathbf{P}(t_b| \underbrace{\vec{\theta}^{tp}_{\omega}}_\text{topical}, \underbrace{\vec{\theta}^{lx}_{\omega}}_\text{lexical}, \underbrace{\vec{\theta}^{\cal{C}(t_b)}_{s_a}}_\text{contextual}) &=  log \; \mathbf{P}(\vec{w}^{t_b}_{out}|\vec{\theta}^{t_b}_{in}) \\
&\approx log \; f(\vec{w}^{t_b}_{out},\vec{\theta}^{t_b}_{in}) + \sum^{k}_{i=1} \mathbb{E}_{t \backsim P_n(t_b)} \big( \llbracket t \neq t_b\rrbracket log \; f(-1 \times \vec{w}^{t}_{out}, \vec{\theta}^{t_b}_{in})\big) \\
f(\vec{w}^{t}_{out},\vec{\theta}^{t_b}_{in}) &= Uh((\vec{w}^{t}_{out})^T \times \vec{\theta}^{t_b}_{in})
\end{split}
\end{equation}	

The negative sampling algorithm tries to distinguish the correct guess $t_b$ with $k$ randomly selected negative samples $\{t | t \neq t_b\}$ using $k+1$ logistic regressions. $\mathbb{E}_{t \backsim P_n(t)}$ is a sampling function that samples a token $v$ from the vocabulary $V$ according to the noise distribution $P_n(t)$ of $V$. By substituting the log probability in Equation~\ref{equ:lexical:j1} with Equation~\ref{equ:lexical:p2} and taking derivatives, respectively, on $\vec{w}^{t}_{out}$ and $\vec{\theta}^{t_b}_{in}$, we have the gradients to be updated:

\begin{equation}
\label{equ:lexical:p2:derivative}
\begin{split}
\frac{\partial }{\partial \vec{w}^{t}_{out}} J(\theta)_1^{ng} &= \bigg(\llbracket t == t_b \rrbracket - f(\vec{w}^{t}_{out},\vec{\theta}^{t_b}_{in})\bigg) \times \vec{\theta}^{t_b}_{in} \\
\frac{\partial }{\partial\vec{\theta}^{t_b}_{in}} J(\theta)_1^{ng} &= \sum^{k}_{i} \mathbb{E}_{t \backsim P_n(t)} \bigg(\big( \llbracket t == t_b \rrbracket - f(\vec{w}^{t}_{out},\vec{\theta}^{t_b}_{in}) \big) \times \vec{w}^{t}_{out} \bigg)
\end{split}
\end{equation}	
The superscript $ng$ of $J(\theta)_1^{ng}$ indicates that the log probability of the objective function is substituted by negative sampling rather than a complete soft-max. For each $t_b$ to be predicted, we only need to update $d_1 \times (k + 1 + \cal{W}(tp) \times 2)$ parameters for the second output. $k$ is contributed by the negative sampling. $\cal{W}(tp) \times 2$ is contributed by the contextual input. We equally propagate the error to each lexical token in the context $\cal{C}(t_b, s_a)$ of $t_b$. The constant 1 is contributed by the lexical bias term $\vec{\theta}^{lx}_{\omega}$. We do not propagate the errors from the first output to the topical modality $\vec{\theta}^{tp}_{\omega}$ since the topical bias is determined by the holistic distribution of vocabulary and is not determined by the specific token selection on the local level. We update $\vec{\theta}^{tp}_{\omega}$ in the second output, which will be described below.

\begin{example}
Continue from Example~\ref{example:topical-lexical}. This example shows how to train the model for the first output. We map $t_4$ into its output vector $\vec{w}^{t_4}_{out}$. Next we can calculate $\mathbf{P}(\vec{w}^{t_4}_{out}|\vec{\theta}^{t_4}_{in})$ using negative sampling (Equation~\ref{equ:lexical:p2}). After that we can calculate the gradients w.r.p.t $\vec{w}^{t_4}_{out}$ and $\vec{\theta}^{t_4}_{in}$ using Equation~\ref{equ:lexical:p2:derivative}. We update $\vec{w}^{t_4}_{out}$ according to its gradient with a pre-specified learning rate. We also update $\vec{w}^{t_2}_{in}$, $\vec{w}^{t_3}_{in}$, $\vec{w}^{t_5}_{in}$,  $\vec{w}^{t_6}_{in}$, and $\vec{\theta}^{lx}_{\omega}$ equally according to the gradient of $\vec{\theta}^{t_4}_{in}$. $\blacksquare$
\end{example}

The second output of this model captures the topical bias reflected on the document $\omega$. The topics reflected from the text can be interpreted as the union of effects of all the local context in the sentence. Thus, the second output of this single-layer feed-forward neural network (see the left part of Figure~\ref{fig:topic}) is a multi-class prediction of each word in the sentence $s_a$, which is denoted by $\cal{T}(s_a)$ in Definition~\ref{def:pro:stylometricRepresentationLearning}. The goal is to maximumize the log probability on $\vec{\theta}^{tp}_{\omega}$ of document $\omega$ for each of its sentences $\cal{S}(\omega)$:

\begin{align}
\label{equ:topical:j1}
\argmax J_2(\theta) = \argmax \frac{1}{|\mathbb{D}|}\sum^{\mathbb{D}}_{\omega}\sum^{\cal{S}(\omega)}_{s_a}\sum^{\cal{T}(s_a)}_{t_b} log \; \mathbf{P}(t_b| \underbrace{\vec{\theta}^{tp}_{\omega}}_\text{topical})
\end{align}

Similar to the first output of this model, we map each lexical token at the output to a numeric vector $\vec{w}^{t_b}_{out}$ (the yellow rectangles in Figure~\ref{fig:topic}). Suppose that we use the typical soft-max multi-class output layer. The second output of this model captures the probability of picking a word $t_b$ based on the topics $\theta^{t_b}_{in}$ as follows:

\begin{equation}
\begin{split}
\label{equ:topical:p1}
\mathbf{P}(t_b| \underbrace{\vec{\theta}^{tp}_{\omega}}_\text{topical}) = \mathbf{P}(\vec{w}^{t_b}_{out} | \vec{\theta}^{tp}_{\omega}) &= \frac{f(\vec{w}^{t_b}_{out}, \vec{\theta}^{tp}_{\omega})}{\sum_{t}^{V}f(\vec{w}^{t}_{out}, \vec{\theta}^{t_b}_{\omega})} \\
f(\vec{w}^{t}_{out},\vec{\theta}^{tp}_{\omega}) &= Uh((\vec{w}^{t}_{out})^T \times \vec{\theta}^{tp}_{\omega})
\end{split}
\end{equation}

The total number of parameters to be estimated is $(|V| + 1) \times d_1$. However, the term $|V|$ is too large. Similar to the first output of this model, we use the $k$ negative sampling approach to approximate the log probability:

\begin{equation}
\begin{split}
\label{equ:topical:p2}
\mathbf{P}(t_b| \underbrace{\vec{\theta}^{tp}_{\omega}}_\text{topical}) &= \mathbf{P}(\vec{w}^{t_b}_{out} | \vec{\theta}^{tp}_{\omega})  \\
&\approx log \; f(\vec{w}^{t_b}_{out},\vec{\theta}^{tp}_{\omega}) + \sum^{k}_{i=1} \mathbb{E}_{t \backsim P_n(t_b)} \big( \llbracket t \neq t_b\rrbracket log \; f(-1 \times \vec{w}^{t}_{out}, \vec{\theta}^{tp}_{\omega})\big)
\end{split}
\end{equation}

By substituting the log probability in Equation~\ref{equ:topical:j1} with Equation~\ref{equ:topical:p2}, and taking the derivatives respectively over $\vec{w}^{t_b}_{out}$ and $ \vec{\theta}^{tp}_{\omega}$, we have the derivatives to be updated for each $t_b$:

\begin{equation}
\label{equ:topical:p2:derivative}
\begin{split}
\frac{\partial }{\partial \vec{w}^{t}_{out}} J(\theta)_2^{ng} &= \bigg(\llbracket t == t_b \rrbracket - f(\vec{w}^{t}_{out},\vec{\theta}^{tp}_{\omega})\bigg) \times \vec{\theta}^{tp}_{\omega} \\
\frac{\partial }{\partial\vec{\theta}^{tp}_{\omega}} J(\theta)_2^{ng} &= \sum^{k}_{i} \mathbb{E}_{t \backsim P_n(t)} \bigg(\big( \llbracket t == t_b \rrbracket - f(\vec{w}^{t}_{out},\vec{\theta}^{tp}_{\omega}) \big) \times \vec{w}^{t}_{out} \bigg)
\end{split}
\end{equation}	

The total number of parameters now becomes $(k + 1) \times d_1$ for each $t_b$. Constant $k$ is contributed by $k$ negative samples, and constant 1 is contributed by the update of $\vec{\theta}^{tp}_{\omega}$. Basically, the second output of this model is an approximation to the full factorization of the document-term co-occurrence matrix.

\begin{example}
Continue from Example~\ref{example:topical-lexical}. This example shows how to train the model for the second output. For the second output of the model we map each token into a numeric vector $\vec{w}^{t_b}_{out}$, where $t_b \in \{$`it', `is', `a', `great', `day', `!!'$\}$. For each of the vectors we calculate $\mathbf{P}(\vec{w}^{t_b}_{out} | \vec{\theta}^{tp}_{\omega})$ in Equation~\ref{equ:topical:p2} using negative sampling. Then we calculate the derivatives for each $\vec{w}^{t_b}_{out}$ and $\vec{\theta}^{tp}_{\omega}$ by using Equation~\ref{equ:topical:p2:derivative}, and update them accordingly by multiplying the gradients with a pre-specified learning rate.~$\blacksquare$
\end{example}

In this model, we count punctuation marks as lexical tokens. Consequently, the information related to the punctuation marks is also included. Punctuation marks carry information of intonation in linguistics and are useful for authorship analysis~\cite{DBLP:journals/jasis/TorneyVY12}. After training the model on a given text dataset $\mathbb{D}$, we have a topical modality vector representation $\vec{\theta}^{tp}_{\omega} \in \mathbb{R}^{d_1}$ and a lexical modality vector representation $\vec{\theta}^{lx}_{\omega} \in \mathbb{R}^{d_1}$ for each document $\omega \in \mathbb{D}$. Also, for each lexical token $t_b \in V$ we have a vectorized representation $\vec{w}^{t_b}_{in} \in \mathbb{R}^{d_1}$.

For an unseen document $\omega^{'} \notin \mathbb{D}$ that does not belong to the training text data, we fix all the $\vec{w}^{t_b}_{in} \in \mathbb{R}^{d_1}$ and $\vec{w}^{t_b}_{out} \in \mathbb{R}^{d_1}$ in the trained model and only propagate errors to $\vec{\theta}^{lx}_{\omega^{'}} \in \mathbb{R}^{d_1}$ and $\vec{\theta}^{tp}_{\omega^{'}} \in \mathbb{R}^{d_1}$. At the end, we also have both $\vec{\theta}^{lx}_{\omega^{'}}$ and $\vec{\theta}^{d_1}_{\omega^{'}}$ for $\omega^{'}$.

The \emph{PVCBOW} model and \emph{PVDM} model for paragraph embedding in \cite{DBLP:journals/corr/LeM14} are trained in a similar way, and all of them operate on the lexical tokens. However, the proposed joint network structure in this paper is customized for authorship analysis. We jointly model the topical bias and the local contextual bias in a single neural network. Thus, it is very different from the \emph{PVCBOW} and \emph{PVDM} models.

\subsection{The character-level modality}
\label{sec:character}

Features of the character-level modality are concerned with the morphology and phonemes biases in the process of constructing/spelling a single lexical word~\cite{DBLP:journals/jasis/TorneyVY12}. The typical feature used for character modality is the character $n$-gram feature. Character $n$-gram is a sequence of consecutive characters. Usually the character bigrams, trigrams, and four-grams (i.e., $n \in {2,3,4}$) with their respective frequency or \emph{tf$\times$idf} values are used for AA studies. They have been shown to be effective for authorship verification~\cite{pan:halvanivebav,DBLP:conf/setn/PothaS14}, attribution~\cite{DBLP:conf/coling/NasirGB14}, and characterization~\cite{DBLP:conf/emnlp/BurgerHKZ11}. \citeN{DBLP:conf/coling/SapkotaSMBR14} show that character $n$-gram features are robust and perform well even in the condition where the training data and testing data are on different topics. In contrast, our previous studies~\cite{ding2015visualizable} show that the lexical $n$-gram approach still outperforms the character $n$-gram approach even on e-mail data where topics vary across different documents. This is possibly because the documents used by~\citeN{DBLP:conf/coling/SapkotaSMBR14} are newspapers, which on average are more formal and longer than e-mail data; therefore, the cumulative effect of character $n$-gram is more stable and robust.

However, similar to the lexical $n$-gram, the problems related to the character $n$-gram are two-fold. First it is difficult to determine the parameter $n$, and the choice actually depends on the data. For formal writings such as theses, academic papers, or newspapers, bigrams and trigrams appear to be sufficient; however, for informal writing such as tweets, some special grams are ignored if only $n \in {2,3,4}$ is considered. For example, a lexical token containing repeated alphabets (e.g., ``niceeeeee") are popular on social networks and they are important for authorship analysis as a socio-linguistic feature~\cite{rao2010detecting}. However using only short $n$-grams cannot really distinguish the style of ``niceeeeee" from the style of the repeating lexical tokens in ``engineer" and ``IEEE" because all of them reflect the frequent usage of bigram ``ee". The question is: Can we model the relationship between characters directly from the text instead of choosing the parameter $n$? This example illustrates that the character modality has a strong relationship with the lexical token. Can we also model the variation of the morphological relationship between them? Second, the number of possible character $n$-grams is large and it is difficult to determine the choice of feature selection method and the hyper parameter $k$. They have a strong dependency on the available dataset, as shown in our experiment. The question is: Can we get rid of these dataset-dependent configurations?

\begin{figure*}
\centering
\includegraphics[width=5.0in]{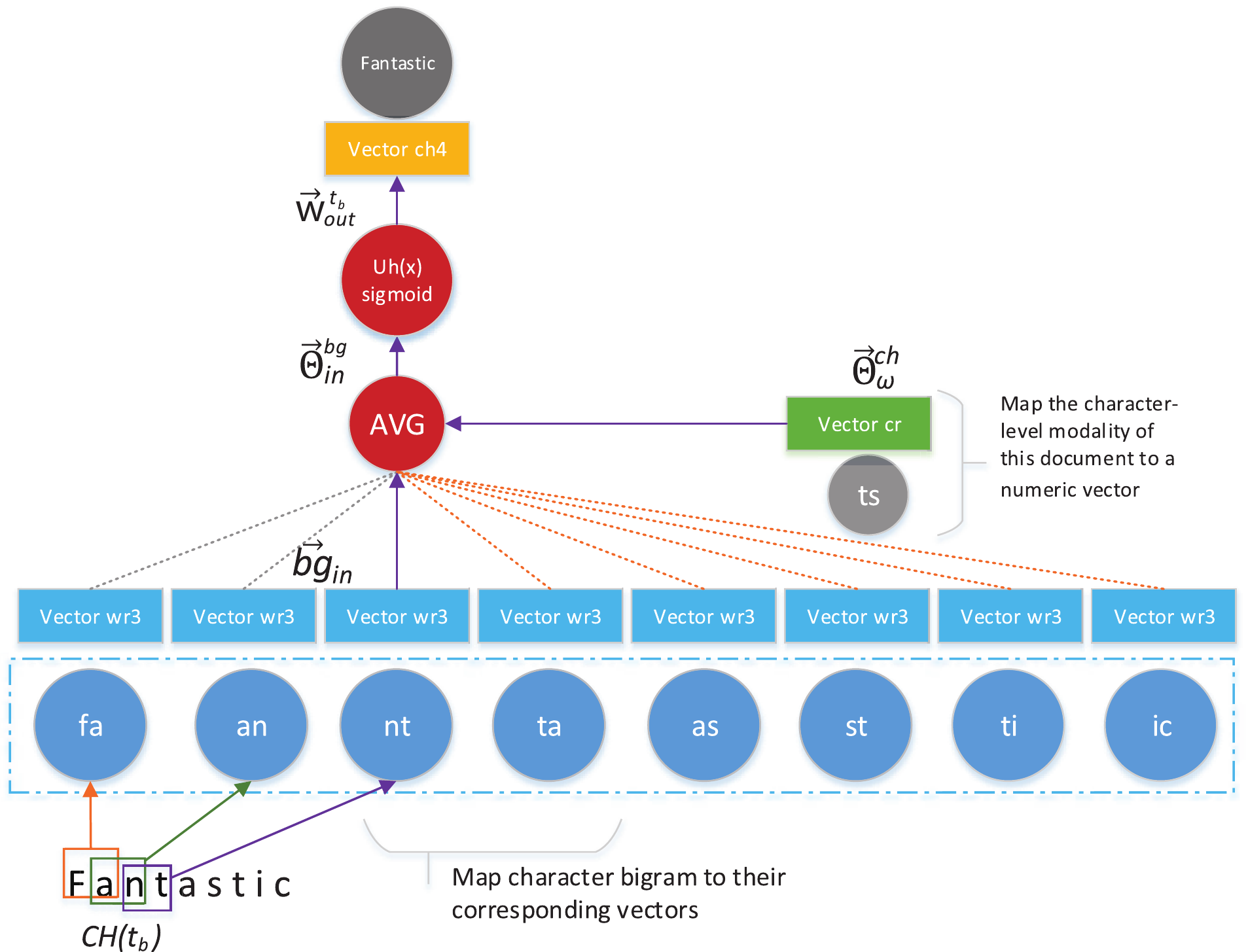}
 
\caption{The model for learning the stylometric representation of the character-level modality.}
\label{fig:character}
\end{figure*}

In order to address the aforementioned issues of character-based stylometric features, we propose a neural-network-based model to learn the character modality representation on the plain text data. This model consists of a single neural layer and captures the morphological differences in constructing and spelling lexical tokens across different documents. Refer to Figure~\ref{fig:character}. The input of this model is one of the character bigrams generated by a sliding window over a lexical token $t_b$ with the character-level bias. The output of this model is the vectorized representation of the token $t_b$. The purpose is to learn $\vec{\theta}^{ch}_{\omega} \in \mathbb{R}^{\cal{D}(ch)}$ for each document $\omega \in \mathbb{D}$ such that vector $\vec{\theta}^{ch}_{\omega}$ captures the morphological differences in constructing lexical tokens. Let $\cal{CH}(t_b) = bg[1:c]$ denote the list of character bigrams of a given token $t_b$, and $bg$ is one of them. The goal is to maximize the following log probability on the given dataset $\mathbb{D}$:

\begin{align}
\label{equ:char:j1}
\argmax J(\theta) = \argmax \frac{1}{|\mathbb{D}|}\sum^{\mathbb{D}}_{\omega}\sum^{\cal{S}(\omega)}_{s_a}\sum^{\cal{T}(s_a)}_{t_b}\sum^{\cal{CH}(t_b)}_{bg} log \; \mathbf{P}(t_b| \underbrace{\vec{\theta}^{ch}_{\omega}}_\text{char-level},\;\vec{bg}_{in})
\end{align}

We use a character bigram instead of unigram to increase the character-level vocabulary size. Similar to the previous lexical model, we map each lexical token $t_b$ into a numeric vector $\vec{w}^{b_t}_{out}$, which is used to output a multi-class prediction.  We also map each character bigram into a numeric vector $\vec{bg}_{in}$, which is used for the network input. Both are model parameters to be estimated. The input vectors of this model are $\vec{bg}^{b_t}_{in}$ and $\vec{\theta}^{ch}_{\omega}$. Both of them have the same dimensionality $d_2$. After taking an average, it is fed into the neural network, as depicted in Figure~\ref{fig:character}, to predict its corresponding lexical token $t_b$. We considered adding a soft-max layer to predict $t_b$:
\begin{equation}
\begin{split}
\label{equ:char:p1}
\vec{\theta}^{bg}_{in} &= \mean{\vec{\theta}^{ch}_{\omega}, \vec{bg}_{in}} \\
\mathbf{P}(t_b| \underbrace{\vec{\theta}^{ch}_{\omega}}_\text{char-level},\vec{bg}_{in}) & = \mathbf{P}(\vec{w}^{t_b}_{out} |\vec{\theta}^{bg}_{in}) = \frac{f(\vec{w}^{t_b}_{out},  \vec{\theta}^{bg}_{in})}{\sum_{t}^{V}f(\vec{w}^{t}_{out}, \vec{\theta}^{bg}_{in})} \\
f(\vec{w}^{t}_{out},\vec{\theta}^{bg}_{in}) &= Uh((\vec{w}^{t}_{out})^T \times \vec{\theta}^{bg}_{in})
\end{split}
\end{equation}
Again, there are $O(V)$ parameters to be updated for each pass of the neural network, which is not efficient. Thus, we use the negative sampling approach to approximate the log probability:
\begin{equation}
\begin{split}
\label{equ:char:p2}
\mathbf{P}(t_b| \underbrace{\vec{\theta}^{ch}_{\omega}}_\text{char-level},bg) & = \mathbf{P}(\vec{w}^{t_b}_{out} |\vec{\theta}^{bg}_{in})   \\
&\approx log \; f(\vec{w}^{t_b}_{out},\vec{\theta}^{bg}_{in}) + \sum^{k}_{i=1} \mathbb{E}_{t \backsim P_n(t_b)} \big( \llbracket t \neq t_b\rrbracket log \; f(-1 \times \vec{w}^{t}_{out}, \vec{\theta}^{bg}_{in})\big)
\end{split}
\end{equation}
Similar to the previous model, we have the following derivatives by using negative sampling as indicated by the superscript $ng$:
\begin{equation}
\label{equ:char:p2:derivative}
\begin{split}
\frac{\partial }{\partial \vec{w}^{t}_{out}} J(\theta)^{ng} &= \bigg(\llbracket t == t_b \rrbracket - f(\vec{w}^{t}_{out},\vec{\theta}^{bg}_{in})\bigg) \times \vec{\theta}^{bg}_{in}) \\
\frac{\partial }{\partial\vec{\theta}^{bg}_{in}} J(\theta)^{ng} &= \sum^{k}_{i} \mathbb{E}_{t \backsim P_n(t)} \bigg(\big( \llbracket t == t_b \rrbracket - f(\vec{w}^{t}_{out},\vec{\theta}^{bg}_{in}) \big) \times \vec{w}^{t}_{out} \bigg)
\end{split}
\end{equation}	

The number of parameters to be updated for each bigram $bg$ of token $t_b$ is $(k+2) \times d_2$. The constant $k$ is contributed by the negative sampling function, and the constant 2 is contributed by $\vec{\theta}^{ch}_{\omega}$ and $\vec{bg}_{in}$. To learn $\vec{\theta}^{ch}_{\omega^{'}}$, for $\omega^{'} \notin \mathbb{D}$ we fix all $\vec{w}^{t_b}_{out}$ and $\vec{bg}_{in}$ and only propagate errors to $\vec{\theta}^{ch}_{\omega^{'}}$.

\begin{example}
\label{example:char}
Consider a simple sentence: $t_a = $ ``Fantastic day !!" in Figure~\ref{fig:character}. For each token $\{t_b|b \in [1,3]\}$ we extract its character bigrams. Suppose the word in the target is $t_1 =$ `fantastic', and its bigrams are $\cal{CH}(t_4) = \{bg_c|c \in {1,2,3,4,5,6,7,8}\} =  \{$`fa', `an', `nt', `ta', `as', `st', `ti', `ic'$\}$. The training process is the same for other words in this sentence. Let us take a bigram $bg_1=$`fa' as an example. First, we map $bg_1$ to its vectorized representation $\vec{bg}_{in}$ and map $t_1$ to its representation $\vec{w}^{t_1}_{out}$. In combination with $\vec{\theta}^{ch}_{\omega}$, we calculate $\vec{\theta}^{bg}_{in}$ according to the first formula in Equation~\ref{equ:char:p1}. Then we calculate the forward log probability for $\mathbf{P}(\vec{w}^{t_1}_{out} |\vec{\theta}^{bg}_{in})$ in Equation~\ref{equ:char:p2}. Then we calculate the corresponding gradients in Equation~\ref{equ:char:p2:derivative} and update the respective parameters. The training pass for bigram $bg_1=$ `fa' is completed, and we move to the next bigram $bg_2$=`an' following the sample procedure. After all the bigrams are completed, we move to the next token $t_2=$  `day'.~$\blacksquare$
\end{example}

\subsection{The syntactic modality}
\label{sec:syntactic}

Syntactic features are usually considered as deep linguistic features that are comparatively more difficult to consciously manipulate~\cite{DBLP:conf/coling/Gamon04}. Typically, the Part-of-Speech (POS) tags $n$-grams are used as the features for this modality~\cite{DBLP:conf/slsp/BoukhaledG14,DBLP:conf/kes/Baron14,DBLP:conf/sigir/QianL0H14}. Given a token $t_b$, its Part-of-Speech (POS) tag represents its grammatical role in the sentence. The number of total unique tags is much smaller than the number of the character $n$-grams and the number of the lexical $n$-grams. Due to its limited feature space, it is less effective than the other linguistic modalities; however, it is more robust than the others~\cite{ding2015visualizable}, especially for the cross-language authorship attribution problem~\cite{bogdanova2014cross}. However, by re-constructing the POS $n$-grams based on the traversing order of the parsed tree structure of the sentence, \cite{DBLP:conf/micai/Posadas-DuranSB14} show that the complete POS tag or \emph{Syntatic Relation} (SR) tag $n$-grams can actually outperform the character $n$-grams when given the same number of features. However, based on their experimental result, we notice that as the limited number of features increases, the performance of character $n$-grams increases steadily without reaching the peak value, and their maximum limit on the number of features appears to limit the performance of the character $n$-gram approach. Recently, \citeN{DBLP:journals/lalc/FengH14} present the syntactic information in a different way by considering the sequence of grammatical roles of the entities recognized from the text, e.g., the entity is the subject, the object, or neither. Another alternative to the POS tags $n$-gram is the fully parsed POS dependent tree of the sentence~\cite{DBLP:conf/eacl/TschuggnallS14,DBLP:conf/micai/Posadas-DuranSB14}. Syntactic tree-based approaches have larger feature space than POS $n$-grams. However, we find that for most online casual text snippets and informal writings the accuracy of such a parsing approach drops significantly. Moreover, it is inefficient and computationally infeasible to parse a large dataset into full syntactic trees.

\begin{figure*}
\centering
\includegraphics[width=5.0in]{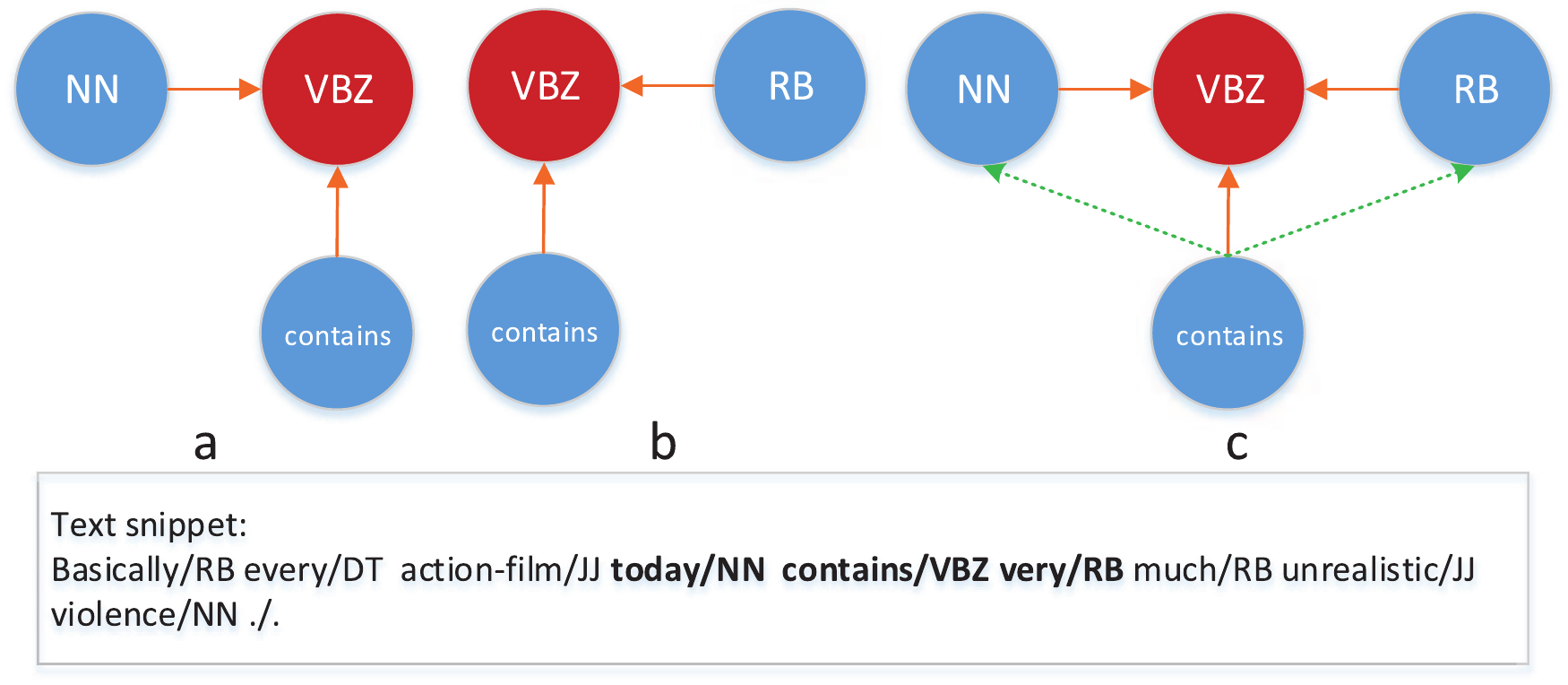}
 
\caption{Three typical inference structures for the Part-Of-Speech tagger.}
\label{fig:syntactic}
\end{figure*}

Instead of looking at the POS $n$-grams, we seek another alternative to maximize the degree of variations that we can gain from the POS tags. First, we look into the state-of-the-art tagger models. POS taggers are pre-trained models that take a list of tokens as input and output a list of tags. Suppose we have a sentence $s_a$ with its tokens $t_b \in \cal{T}(s_a)$. Recall that $\cal{P}(t_b)$ denotes the POS tag for the token $t_b$ in the sentence. Refer to Definition~\ref{def:pro:stylometricRepresentationLearning}. To assign a tag $\cal{P}(t_b)$ to a token $t_b$, there are three typical structures/models~\cite{ref:pos}:
\begin{itemize}
\item
\emph{Left-to-Right structure}. This structure tries to maximize $\mathbf{P}(\cal{P}(t_b)|t_b, \cal{P}(t_{b-1}))$. The tag for token $t_b$ is determined by both the lexical token itself and the next tag $\cal{P}(t_{b-1})$. Strong dependencies exist between $\cal{P}(t_{b-1})$ and $\cal{P}(t_{b})$ and between $\cal{P}(t_{b})$ and $t_b$. See Figure~\ref{fig:syntactic}a. The solid lines indicate the dependencies.
\item
\emph{Right-to-Left structure}. This structure tries to maximize $\mathbf{P}(\cal{P}(t_b)|t_b, \cal{P}(t_{b+1}))$. The tag for token $t_b$ is determined by both the lexical token itself and the previous tag $\cal{P}(t_{b+1})$. Strong dependencies exist between $\cal{P}(t_{b+1})$ and $\cal{P}(t_{b})$ and between $\cal{P}(t_{b})$ and $t_b$. See Figure~\ref{fig:syntactic}b. The solid lines indicate the dependencies.
\item
\emph{Bidirectional structure} This structure is a combination of the previous two. It maximizes $\mathbf{P}(\cal{P}(t_b)|t_b, \cal{P}(t_{b+1}), \cal{P}(t_{b-1}))$. The tag for token $t_b$ is determined by both the lexical token itself and the surrounding tags $\cal{P}(t_{b+1})$ and $\cal{P}(t_{b-1})$. Strong dependencies exist between $\cal{P}(t_{b+1})$ and $\cal{P}(t_{b})$, between $\cal{P}(t_{b})$ and $\cal{P}(t_{b-1})$, and between $\cal{P}(t_{b})$ and $t_b$. See Figure~\ref{fig:syntactic}c. The solid lines indicate the dependencies.
\end{itemize}

For all of these three structures, there exists a strong dependency between contiguous POS tags, as well as between the actual lexical token and its tag. Using POS tags $n$-grams as a stylometric feature is less effective than using character $n$-grams and lexical $n$-grams because the strong dependencies between contiguous POS tags introduced by the POS taggers are shared between different documents. An $n$-gram-based model enlarges the feature space using the contiguous gram dependencies, while for POS $n$-grams it is weakened by the taggers.

Therefore, we seek another way that has fewer dependencies introduced by the POS tagger. In Figure~\ref{fig:syntactic}c, strong dependencies introduced by the tagger are shown as solid lines. We select two weak dependency links from $t_b$ to $\cal{P}(t_{b+1})$ and from $t_b$ to $\cal{P}(t_{b-1})$, as indicated by the dashed lines. The tagger only introduces indirect dependencies on these two paths. Thus, these two paths have more variations across different documents than the others, as indicated by solid lines. Formally, our model tries to maximize $P(\cal{P}(t_{b-1}),\cal{P}(t_{b+1})|t_b)$, which is different from the typical structures for the taggers.

\begin{figure*}
\centering
 
\includegraphics[width=5.0in]{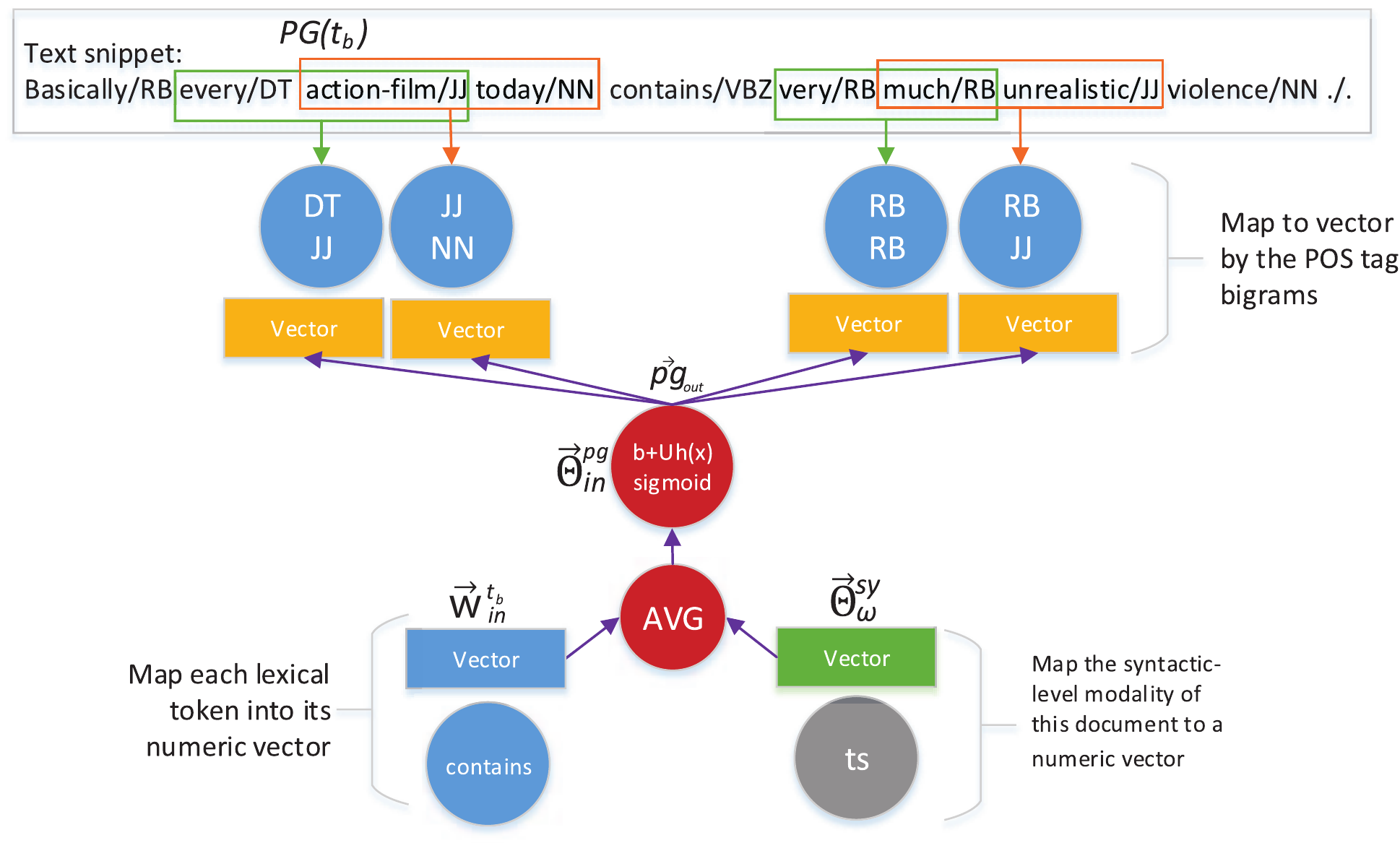}
\caption{The model for learning the stylometric representation of the syntactic modality.}
\label{fig:syntactic2}
\end{figure*}

The number of unique POS tags is quite limited, so we use the bigrams of POS tags. See Figure~\ref{fig:syntactic2}. Let $\cal{P}_2(t_b)$ be a POS tag bigram $[\cal{P}(t_{b}),\cal{P}(t_{b+1})]$, and $pg^b = \cal{PG}(t_{b}) = \{\cal{P}_2(t_{b-3}), \cal{P}_2(t_{b-2}), \cal{P}_2(t_{b+1}), \cal{P}_2(t_{b+2})\}$ be the neighbor POS bigrams of token $t_b$. The goal of this model is to maximize:

\begin{align}
\label{equ:syntactic:j1}
\argmax J(\theta) = \argmax \frac{1}{|\mathbb{D}|}\sum^{\mathbb{D}}_{\omega}\sum^{\cal{S}(\omega)}_{s_a}\sum^{\cal{T}(s_a)}_{t_b}\sum^{\cal{PG}(t_{b})}_{pg^b} log \; \mathbf{P}(pg^b| \underbrace{\vec{\theta}^{sy}_{\omega}}_\text{syntactic},\; \vec{w}^{t_b}_{in})
\end{align}

Similar to the previous models, this model maps each lexical token $t_b$ into a numeric vector $\vec{w}^{t_b}_{in}$, and each of its neighbor POS bigrams maps into an numeric vector $\vec{pg}^b_{out}$. The input of the model, denoted by $\vec{\theta}^{pg}_{in} $, is the average of $\vec{w}^{t_b}_{in}$ and $\vec{\theta}^{sy}_{\omega}$, and the prediction is one of the target token $t_b$'s neighbor POS tag bigrams, as shown in Figure~\ref{fig:syntactic2}. $\vec{w}^{t_b}_{in}$ and $\vec{\theta}^{sy}_{\omega}$ share the same dimensionality $d_3$. The prediction can be implemented as a soft-max layer:
\begin{equation}
\begin{split}
\label{equ:syntactic:p1}
\vec{\theta}^{pg}_{in} &= \mean{\vec{\theta}^{sy}_{\omega}, \vec{w}^{t_b}_{in}} \\
\mathbf{P}(pg^b| \underbrace{\vec{\theta}^{sy}_{\omega}}_\text{syntactic},t_b) & = \mathbf{P}(\vec{pg}^b_{out} |\vec{\theta}^{pg}_{in}) = \frac{f(\vec{pg}^b_{out},  \vec{\theta}^{pg}_{in})}{\sum_{pg}^{V_{pg}}f(\vec{pg}_{out}, \vec{\theta}^{pg}_{in})} \\
f(\vec{pg}_{out},\vec{\theta}^{pg}_{in})) &= Uh((\vec{pg}_{out})^T \times \vec{\theta}^{pg}_{in}))
\end{split}
\end{equation}
where $V_{pg}$ denotes the union of all distinct POS bigrams, and the number of parameters to be updated for each $pg^b$ is bounded by $V_{pg}$, which is around a few hundreds. It is still computationally feasible to directly use the soft-max layer. It is possible to use the negative sampling as well:
\begin{equation}
\begin{split}
\label{equ:syntactic:p2}
\mathbf{P}(pg^b| \underbrace{\vec{\theta}^{sy}_{\omega}}_\text{syntactic},t_b) & = \mathbf{P}(\vec{pg}^b_{out} | \vec{\theta}^{pg}_{in})   \\
&\approx log \; f(\vec{pg}^b_{out} ,\vec{\theta}^{pg}_{in}) + \sum^{k}_{i=1} \mathbb{E}_{pg \backsim P_n(pg^b)} \big( \llbracket pg \neq pg^b\rrbracket log \; f(-1 \times\vec{pg}_{out} , \vec{\theta}^{pg}_{in})\big)
\end{split}
\end{equation}
where $P_n(pg^b)$ denotes the negative sampling function for $V_{pg}$. Accordingly, we have the following derivatives for back propagation:

\begin{equation}
\label{equ:syntactic:p2:derivative}
\begin{split}
\frac{\partial }{\partial \vec{pg}^b_{out}} J(\theta)^{ng} &= \bigg(\llbracket pg == pg^b \rrbracket - f(\vec{pg}^b_{out},\vec{\theta}^{pg}_{in})\bigg) \times \vec{\theta}^{pg}_{in}) \\
\frac{\partial }{\partial\vec{\theta}^{pg}_{in}} J(\theta)^{ng} &= \sum^{k}_{i} \mathbb{E}_{pg \backsim P_n(pg^b)} \bigg(\big( \llbracket pg == pg^b \rrbracket - f(\vec{pg}_{out},\vec{\theta}^{pg}_{in}) \big) \times \vec{pg}_{out} \bigg)
\end{split}
\end{equation}	
At the end of the training, we have $\vec{\theta}^{sy}_{\omega}$ for each document $\omega \in \mathbb{D}$. To estimate $\vec{\theta}^{sy}_{\omega^{'}}$  for $\omega^{'} \notin \mathbb{D}$, we fix all $\vec{w}^{t_b}_{in}$ and $\vec{pg}_{out}$ and only propagate errors to $\vec{\theta}^{sy}_{\omega^{'}}$.

\begin{example}
\label{example:syntactic}
Consider a simple sentence and its corresponding sequence of POS tags in Figure~\ref{fig:syntactic2}. For each token $\{t_b|b \in [1,10]\}$ we extract its POS neighbor bigrams. Suppose the word in target is $t_5 =$`contains', and its POS neighbor bigrams are $\cal{PG}(t_5) = \{$`DT JJ', `JJ NN', `RB RB', `RB JJ'$\}$ given a window size of 2. The training process is the same for other lexical tokens in the sentence. Let us take one of its ($t_5$'s) POS neighbor bigrams $pg^5=$`DT JJ' as an example. First we map $pg^5$ to its vectorized representation $\vec{pg}^5_{in}$ and map $t_5$ to its representation $\vec{w}^{t_5}_{in}$. With $\vec{\theta}^{sy}_{\omega}$, we calculate $\vec{\theta}^{pg}_{in}$ according to the first formula in Equation~\ref{equ:syntactic:p1}. In combination with $\vec{pg}^5_{in}$, we calculate the forward log probability for $\mathbf{P}(\vec{pg}^5_{in} |\vec{\theta}^{pg}_{in})$ in Equation~\ref{equ:syntactic:p2}. Then we calculate the corresponding gradients in Equation~\ref{equ:syntactic:p2:derivative} and update the respective parameters. The training pass for bigram $pg^5=$'DT JJ' is completed, and we move to the next bigram `JJ NN' following the same procedure. After all the bigrams are processed, we move to the next token $t_6$.~$\blacksquare$
\end{example}

\subsection{Making the model deterministic}
\label{sec:deterministic}
Deterministic and reproducible results are important requirements for most authorship applications, especially in the area of cyber forensic and linguistic evidence~\cite{DBLP:journals/isci/IqbalBFD13,ding2015visualizable}. We could not show that a piece of document is written by an author in the first run, and later show that it is authored by another author in the subsequent runs with the same inputs and settings. Unfortunately, the proposed stylometric representation learning approach is based on the stochastic gradient descend back-propagation algorithm, which involves a high degree of randomness in its nature. In order to make the proposed model deterministic, we need to enforce specific modifications to the aforementioned models on the implementation level:
\begin{longitem}
\item \emph{Initializing the parameters to be estimated}. To start with, we need to initialize all the neural network input layer parameters of all the models to small random values around zero. In order to have the same sequence of the random number generated for the parameters, we fix the starting random seed as 0. In this way, we make sure that the gradient descend process always starts from the same point at the feature space.

\item \emph{Using the single-thread implementation}. A multi-threaded implementation of the stochastic gradient descent can greatly reduce the runtime for learning stylometric representation; however, the behavior of each thread can hardly be controlled. Even the parameter space is sparse; there is a high chance for different threads to update the parameters of the frequent terms at the same time. To avoid the unpredictable behavior of multi-threading, we choose a single-threaded implementation for both the parameter initialization and stochastic gradient descent. Thus, using the negative sampling approach is critical for all the models to keep the training efficient.
\end{longitem}

\section{Evaluation on Unsupervised Authorship Verification Problem}
\label{sec:verification}

In this section, we evaluate the proposed models on the authorship verification problem. The problem is to verify whether or not two anonymous text documents are written by the same author. Unlike the authorship identification problem, where a set of candidate authors are available for comparison, the authorship verification problem has only one target document to be compared.  The solution to this problem should provide a confidence value that indicates how likely the two given text documents are written by the same author. Authorship identification is a closed-set classification problem and authorship verification is an open-set classification problem~\cite{DBLP:conf/clef/StamatatosDVSPJSB14}. Authorship verification is more difficult to solve than authorship identification.

We further divide authorship verification into two types: supervised verification and unsupervised verification. In the supervised authorship verification problem, the ground-truth data is available in the training set. The ground-truth data consists of a list of authors with their respective written documents. The ground-truth data has similar properties to the two anonymous documents to be verified. For example, all of them are e-mails. Learning-to-rank-based classification schemes fit into this category: Given a vector, the classification model learns to output a value ranging from 0 to 1 based on the training vectors. Typically, a SVM model or a logistic regression model is employed.

In the unsupervised authorship verification problem, no such ground-truth data is available. However, a list of documents that share similar properties to the two anonymous documents to be verified is available. For example, all of them are e-mails. The authors of this list of documents are unknown. The unsupervised authorship verification problem is more difficult than the supervised authorship verification problem. In this section, we focus on the unsupervised authorship verification problem.

\subsection{The solution}

To solve the targeted problem with our proposed stylometric representation learning models, we first train the three models mentioned in Section~\ref{sec:mining} on the unlabeled text data, and then we estimate the stylometric representations $\vec{\theta}^{tp}_{\omega} \in \mathbb{R}^{d_1}$, $\vec{\theta}^{lx}_{\omega} \in \mathbb{R}^{d_1}$, $\vec{\theta}^{ch}_{\omega} \in \mathbb{R}^{d_2}$, and $\vec{\theta}^{sy}_{\omega} \in \mathbb{R}^{d_3}$, respectively, for the two anonymous documents $\omega_1, \omega_2$ in the testing data that is previously unseen by the models. The verification score is a simple cosine distance measure between the given two documents' stylometric representations. Formally, given two anonymous documents $\omega_1, \omega_2$, the solution outputs the similarity value:
\begin{equation}
\cal{Q}(\omega_1, \omega_2) = \frac{(\vec{\theta}^{\mathit{v}}_{\omega_1})^T \times \vec{\theta}^{\mathit{v}}_{\omega_2} }{|\vec{\theta}^{\mathit{v}}_{\omega_1}| \times |\vec{\theta}^{\mathit{v}}_{\omega_2}|} \;\;\;\;\;\;\;\;\;v \in \{tp, lx, ch, sy\}
\end{equation}
where $\mathit{v}$ denotes the selected modality. It could be $tp$ topical modality, $lx$ lexical modality, $ch$ character-level modality, $sy$ syntactic modality, or their combinations. If more than one modality is selected, we concatenate their $\vec{\theta}^{\mathit{v}}_{\omega}$ into a single one for each $\omega$.

To measure the performance of the proposed approaches and the baselines on this problem, we use the Area Under Receiver Operating Characteristic curve (AUROC)~\cite{AUROC}. It is a well-known evaluation measure for binary classifiers where both positive labels and negative labels are equally important. Since changing the threshold of the similarity value results in different accuracy and recall measures, the AUROC measure captures the overall performance of the classifier when the threshold is varied. A 1.0 AUROC value indicates an excellent performance, and a 0.5 AUROC value implies a worthless random guess.

\subsection{The English novel and essay dataset}
\label{sec:verification:dataset}

We choose the PAN2014 authorship verification English dataset as the benchmark dataset. PAN provides a series of shared tasks on digital text forensics. In this way we can directly compare our results with other studies. The latest available dataset (both training and testing) for the authorship verification problem is from PAN2014\footnote{PAN2014 Authorship Verification. Available at \url{http://pan.webis.de/clef14/pan14-web/author-identification.html}}. At the moment of writing this paper, PAN2015 has only published the training dataset for this problem. Currently, we only focus on English text data, even though the aforementioned models can be adopted for different languages.

\begin{figure*}
\centering
\includegraphics[width=5in]{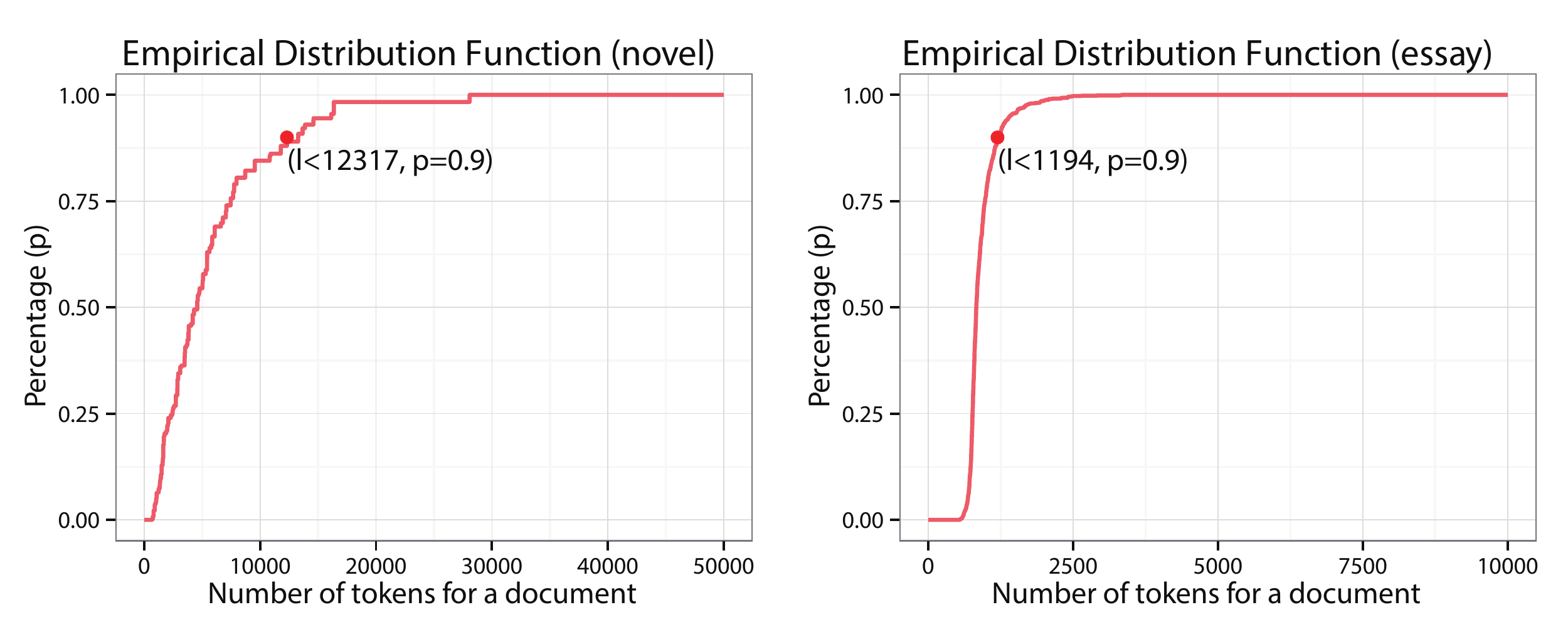}
 
\caption{Empirical Distribution Function on document length in terms of number of tokens for the novel dataset and the essay dataset.}
\label{fig:verification:dataset}
\end{figure*}

\begin{table*}[h]
\centering
\tbl{The PAN2014 authorship verification English dataset.\label{table:verification:dataset}}{
\begin{tabular}{l|rrrrr}
\hline \hline
\begin{tabular}[l]{@{}l@{}}\textbf{}\end{tabular}
 &
\begin{tabular}[l]{@{}l@{}}\textbf{Genre}\end{tabular}
 &
\begin{tabular}[l]{@{}l@{}}\textbf{Verification Cases}\end{tabular}
&
\begin{tabular}[r]{@{}r@{}}\textbf{Documents}\end{tabular}
&
\begin{tabular}[r]{@{}r@{}}\textbf{Sentences}\end{tabular}
&
\begin{tabular}[r]{@{}r@{}}\textbf{Tokens}\end{tabular}
 \\ \hline

\multirow{2}{*}{\begin{tabular}[l]{@{}l@{}}Training Data\end{tabular}}
& English-essays & 200 & 729 & 30,038 & 676,966  \\ \cline{3-6}
& English-novels & 100 & 200 & 28,054 & 705,751  \\ \hline

 \multirow{2}{*}{\begin{tabular}[l]{@{}l@{}}Testing Data\end{tabular}}
& English-essays & 200 & 718 & 29,375 & 653,981  \\ \cline{3-6}
& English-novels & 200 & 400 & 119,202 & 2,781,425  \\ \hline \hline

\end{tabular}
}
\end{table*}

Refer to Table~\ref{table:verification:dataset}. This dataset provides both the training data and testing data. The training data consists of 300 verification cases. Each verification case consists of two sets of documents and a label. The label can be $\mathit{true}$, which indicates that two sets of documents are written by the same author, or $\mathit{false}$, vice versa. 200 of them are essays, and 100 are novels. The test data follows the same format. It contains 400 verification cases; 200 of them are essays, and 200 are novels. Table~\ref{table:verification:dataset} shows that the numbers of essay verification cases for both training and testing data are comparable, while there are more verification cases in testing data than in training data for novels. Figure~\ref{fig:verification:dataset} shows the empirical distribution function over the document length in terms of the size of lexical tokens for essays and novels. It is apparent that documents in the novel dataset are longer than those in the essay dataset. We expect that the proposed model performs better on the novel dataset than the essay dataset based on our previous studies on the factors that influence the quality of AA results~\cite{ding2015visualizable}.

We preprocess the data by removing extra spaces and non-ASCII lexical tokens. We also pre-tokenize texts, detect sentence boundaries, and generate POS tags for the dataset using the Stanford tagger~\cite{ref:pos}. This tagger has a bidirectional structure, discussed in Section~\ref{sec:syntactic}.

As we focus on the unsupervised authorship verification problem, we treat all the training data as unlabeled data (all the ground-truth labels are stripped for training). Only a small portion of randomly sampled problems from the training dataset with their ground-truth labels are selected as a validation set for tuning the hyper-parameters for the proposed models and all the baseline models.

\subsection{Baselines}

We choose several of the most relevant approaches to compare with our proposed models on the authorship verification problem.
\begin{longitem}
\item\emph{Style}. This approach represents a document as a numeric vector under a typical $302$ static stylometric features, which has been widely studied. Table~\ref{table:baseline:stylometry} provides a summary of these features. It is shown to be effective coupled with classifiers for the authorship identification problem. It is called a static feature set since the features do not change across different datasets. The similarity between two documents is based on their normalized cosine distance.

\item \emph{Style+[$k$-freq-$n$gram]}. This approach represents a document as a numeric vector under the 302 static features in Table~\ref{table:baseline:stylometry} as well as $k$ dynamic features constructed based on the training dataset. The $k$ dynamic features are constructed by picking the top-$k$ $n$-grams ranked by their occurring frequency in the training set. $n$-grams include lexical unigrams, lexical bigrams, lexical trigrams, character unigrams, character bigrams, character trigrams, POS tag unigrams, POS tag bigrams, and POS tag trigrams. In this experiment, we pick $k \in \{100,200,500,1000,2000,5000\}$. The value for each top-$k$ selected $n$-gram is calculated using $tf \times idf$. The similarity between two documents is based on their normalized cosine distance.

\item \emph{Style+[$k$-info-$n$gram]}. This approach is the same as \emph{Style+[$k$-freq-$n$gram]} except that the top-$k$ $n$-grams are selected based on information gain rather than the frequency value in the training dataset. Even though previous AA research already experimentally demonstrates that frequency value carries enough stylistic information and outperforms the information gain scheme~\cite{stamatatos2009survey}, we include it in the baselines. We pick $k \in \{100,200,500,1000,2000,5000\}$.

\item \emph{LDA}. The Latent Dirichlet allocation (LDA) is a generative model that learns a latent semantic representation between the documents and the words. The latent semantic operation is learned through Gibbs Sampling. This approach represents each document as a numeric vector under the document-to-latent-topic distribution learned from the dataset. The numeric vector has $k$ elements and each of them corresponds to one of the latent topics in the LDA model. We validate the number of iterations and the number of $k$ for the LDA model on the validation set. The similarity between two documents is the cosine distance between their vectorized representation. We pick $k=500$, which achieves the best result on the validation set.

\item \emph{LSA}. Latent semantic analysis (LSA) is a technique for analyzing the relationship between words and documents. Given a set of documents, we can represent it as a sparse matrix, where a row denotes a document and a column represents a word. A word is represented as its occurrences in different documents and a document is represented as a set of words with their corresponding occurrence in this document. Given such a sparse matrix, LSA learns the latent representation between document and word by factorizing the sparse matrix using Single Value Decomposition (SVD). After SVD, each document can be represented as weights over the $k$ singular values in SVD. The similarity between two documents is the cosine distance between their vectorized representation. We pick $k=200$ by maximizing its performance on the validation set.

\item \emph{w2v-skipgram}. \emph{w2v-skipgram} is a neural network language model that learns the vectorized embedding for words in a text dataset~\cite{DBLP:journals/corr/abs-1301-3781}. It is efficient and has been adapted in various data mining problems. It is well known for the word analogy task by mathematically manipulating the vectorized representation of words. It is shown that this model is an approximated factorization of the co-occurrence matrix between words. By converting each word of a document into a vector and taking their average, we can obtain an vectorized representation of the document. Finally we use cosine similarity to measure the distance between vectors.

\item \emph{w2v-cbow}. \emph{w2v-cbow} is another neural network language model that learns the vectorized embedding for words in a text dataset~\cite{DBLP:journals/corr/abs-1301-3781}. It is more scalable to larger dataset than the \emph{w2v-skipgram} model. Following the previous \emph{w2v-skipgram} approach, we obtain a vectorized representation of a document by converting each word of the document into a vector and taking their average. At the end we use cosine similarity to measure the distance between vectors.

\newcolumntype{L}[1]{>{\raggedright\let\newline\\\arraybackslash\hspace{0pt}}m{#1}}
\begin{table}
\tbl{Baseline static features.\label{table:baseline:stylometry}}
{
\begin{tabular}{c|ccc}
\hline \hline
\textbf{Feature type} & \textbf{Features} & \textbf{Count} & \textbf{Example}\\ \hline
\multirow{6}{*}{Static feature} & \multicolumn{1}{l}{Lexical features} & \multicolumn{1}{c}{105}
								& \multicolumn{1}{L{43mm}}{Ratio of digits and vocabulary richness, etc.  } \\ \cline{2-4}
								& \multicolumn{1}{l}{Function words} & \multicolumn{1}{c}{150}
								& \multicolumn{1}{L{43mm}}{Occurrence of \textit{after} } \\ \cline{2-4}
								& \multicolumn{1}{l}{Punctuation marks} & \multicolumn{1}{c}{9}
								& \multicolumn{1}{L{43mm}}{Occurrences of punctuation \textit{!} } \\ \cline{2-4}
								& \multicolumn{1}{l}{Structural features} & \multicolumn{1}{c}{15} & \multicolumn{1}{L{43mm}}{Presence/absence of greetings}  \\ \cline{2-4}
								& \multicolumn{1}{l}{Domain-specific features} & \multicolumn{1}{c}{13} & \multicolumn{1}{L{43mm}}{Occurrences of word \textit{contract}, \textit{time}, and \textit{draft}, etc.}  \\ \cline{2-4}
								& \multicolumn{1}{l}{Gender-preferential features} & \multicolumn{1}{c}{10} & \multicolumn{1}{L{43mm}}{Ratio of words ending with \textit{ful}}  \\ \hline \hline
\end{tabular}
}
\end{table}

\item \emph{PVDBOW}. \emph{PVDBOW} is a recently proposed model that learns a vectorized document representation based on the neural network language model~\cite{DBLP:journals/corr/LeM14}. It captures the occurrences relationship between words. It has been shown to be effective in sentiment analysis~\cite{DBLP:journals/corr/LeM14}. Our proposed model uses a similar neural network approach but the whole structure and the motivation are different. We validate hyper-parameters for PVDBOW on the validation set. We pick $k=400$ with sub-sampling enabled and a window size of 4 accordingly to maximize its performance on the validation set.

\item \emph{PVDM}. \emph{PVDM} is another recently proposed model that learns a vectorized document representation based on the neural network language model~\cite{DBLP:journals/corr/LeM14}. It captures the \emph{document-to-word} occurrences relationship. It has been shown to be more effective than \emph{PVDBOW} in sentiment analysis~\cite{DBLP:journals/corr/LeM14}. We pick $k=400$ with sub-sampling enabled and a window size of 4 according to its performance on the validation set.

\item Other approaches are reported in PAN2014. For comparison, we have selected the top 10 approaches from the other studies reported in PAN2014. PAN2014 also has a meta-classifier, called \emph{META-CLASSIFIER-PAN2014}, which combines all the submitted approaches.
\end{longitem}
These baselines cover both the recent development in text embedding learning models and authorship verification solutions. We also include several combinations such as \emph{w2v-cbow+skipgram}.  Following the same procedure for the baseline approaches, we train our proposed three models on the training set and choose its hyper-parameters  based on the validation set. We set $b_1 = 400$, $b_2 = 400$, $b_3 = 400$ and select a window size of 4 for the joint model of lexical and topical modality. Evaluation results are reported based on the performance on the test dataset.

\begin{table*}
\centering
\tbl{Performance comparison for the authorship verification problem on the PAN2014 dataset. Entries with \textbf{*} are the performance of our proposed approaches.\label{table:verification:compare}}{
\begin{tabular}{l|rlll}
\hline \hline
\begin{tabular}[l]{@{}l@{}}\textbf{Measure}\end{tabular}
 &
\begin{tabular}[l]{@{}l@{}}\textbf{Approach}\end{tabular}
 &
\begin{tabular}[l]{@{}l@{}}\textbf{Essay}\end{tabular}
&
\begin{tabular}[r]{@{}r@{}}\textbf{Novel}\end{tabular}
&
\begin{tabular}[r]{@{}r@{}}\textbf{Average}\end{tabular}
 \\ \hline

\multirow{2}{*}{\begin{tabular}[l]{@{}l@{}}\\ \\ \\ AUROC\end{tabular}}
&Modality [Lexical+Topical] \textbf{*}	& \textbf{0.8104}	& 0.7796	& \textbf{0.7950}\\ \cline{3-5}
&Modality [Lexical] \textbf{*}	& 0.8092	& 0.7356	& 0.7724\\ \cline{3-5}
&Modality [Character] \textbf{*}	& 0.7416	& \textbf{0.7876}	& 0.7646\\ \cline{3-5}
&META-CLASSIFIER-PAN2014	& 0.7810	& 0.7320	& 0.7565\\ \cline{3-5}
&PVDM	& 0.7234	& 0.7590	& 0.7412\\ \cline{3-5}
&PVDBOW+PVDM	& 0.7294	& 0.7463	& 0.7379\\ \cline{3-5}
&Modality [Topical] \textbf{*}	& 0.7208	& 0.7356	& 0.7282\\ \cline{3-5}
&PVDBOW	& 0.6818	& 0.7142	& 0.6980\\ \cline{3-5}
&\citeN{pan:satyamstatistical}	& 0.6990	& 0.6570	& 0.6780\\ \cline{3-5}
&\citeN{pan:khonji2014slightly} \& Iraqi	& 0.5990	& 0.7500	& 0.6745\\ \cline{3-5}
&\citeN{pan:frery2014ujm}	& 0.7230	& 0.6120	& 0.6675\\ \cline{3-5}
&\citeN{pan:zamani2014authorship} et al.	& 0.5850	& 0.7330	& 0.6590\\ \cline{3-5}
&\citeN{pan:modaresilanguage}	& 0.6030	& 0.7110	& 0.6570\\ \cline{3-5}
&LSA-k=200	& 0.7172	& 0.5956	& 0.6564\\ \cline{3-5}
&Modality [Syntactic] \textbf{*}	&0.6629	&0.6201	&0.6415\\ \cline{3-5}
&\citeN{pan:mayorsingle}	& 0.5720	& 0.6640	& 0.6180\\ \cline{3-5}
&Moreau et al.	& 0.6200	& 0.5970	& 0.6085\\ \cline{3-5}
&\citeN{pan:halvanivebav}	& 0.6290	& 0.5690	& 0.5990\\ \cline{3-5}
&LDA-k=500	& 0.5608	& 0.6320	& 0.5964\\ \cline{3-5}
&\citeN{pan:castillounsupervised}	& 0.5490	& 0.6280	& 0.5885\\ \cline{3-5}
&Static+[5000-info-ngram]	& 0.4727	& 0.6765	& 0.5746\\ \cline{3-5}
&Static+[2000-info-ngram]	& 0.4760	& 0.6684	& 0.5722\\ \cline{3-5}
&Static+[1500-info-ngram]	& 0.4760	& 0.6650	& 0.5705\\ \cline{3-5}
&Static+[1000-info-ngram]	& 0.4752	& 0.6637	& 0.5695\\ \cline{3-5}
&Static+[0200-info-ngram]	& 0.4752	& 0.6625	& 0.5689\\ \cline{3-5}
&Static+[0100-info-ngram]	& 0.4752	& 0.6621	& 0.5687\\ \cline{3-5}
&Static	& 0.4752	& 0.6617	& 0.5685\\ \cline{3-5}
&Static+[0500-info-ngram]	& 0.4752	& 0.6614	& 0.5683\\ \cline{3-5}
&Static+[0100-freq-ngram]	& 0.4716	& 0.6484	& 0.5600\\ \cline{3-5}
&\citeN{pan:harveyauthor:pos} & 0.5790	& 0.5400	& 0.5595\\ \cline{3-5}
&Static+[0200-freq-ngram]	& 0.4725	& 0.6412	& 0.5569\\ \cline{3-5}
&Static+[0500-freq-ngram]	& 0.4712	& 0.6376	& 0.5544\\ \cline{3-5}
&Static+[5000-freq-ngram]	& 0.4666	& 0.6361	& 0.5514\\ \cline{3-5}
&Static+[1000-freq-ngram]	& 0.4652	& 0.6355	& 0.5504\\ \cline{3-5}
&\citeN{pan:laytonsimple}	& 0.5900	& 0.5100	& 0.5500\\ \cline{3-5}
&Static+[1500-freq-ngram]	& 0.4659	& 0.6338	& 0.5499\\ \cline{3-5}
&Static+[2000-freq-ngram]	& 0.4623	& 0.6332	& 0.5478\\ \cline{3-5}
&w2v-cbow+skipgram	& 0.3936	& 0.6902	& 0.5419\\ \cline{3-5}
&w2v-cbow	& 0.3950	& 0.6717	& 0.5334\\ \cline{3-5}
&w2v-skipgram	& 0.3465	& 0.6616	& 0.5041 \\ \hline \hline
\end{tabular}
}
\end{table*}

\subsection{Performance comparison}

In this section, we present our evaluation result on the English authorship verification dataset with respect to the AUROC measure with all the baselines mentioned above. As indicated in Table~\ref{table:verification:compare}, our proposed \emph{Modality} models achieve the highest AUROC score on this authorship verification problem. Specifically, on average the first-rated model is the joint learning model for lexical modality and the topical modality described in Section~\ref{sec:topic}. This model also outperforms all the others on the essay dataset. The runner-up is the lexical modality representation that is learned in the joint learning model. Character-level modality achieves the highest score on the novel dataset. It also outperforms all the aforementioned baselines on average. The syntactic modality does not perform as well as the lexical, topical, and character-level modalities; however, it still achieves better AUROC than the PVDBOW, LSA, LDA, and other dynamic $n$-gram approaches. It is noted that our proposed syntactic modality representation outperforms the other POS-tags-based approach, such as~\cite{pan:harveyauthor:pos} and $n$-gram approaches, that involve POS tags.

Our proposed models perform better than LSA and LDA approaches, and the LSA approaches outperform the LDA approaches. Probably it is because our model is a joint effect of document-to-word co-occurrence relationship and co-occurrence relationship between words, while LSA and LDA only consider the direct relationship between document and word. The PVDBOW and PVDM approaches also outperform LSA and LDA. In general, the neural-network-based model achieves better performance. Table~\ref{table:verification:compare} also shows that our proposed lexical modality representation outperforms the dynamic $n$-gram-based feature representation with a lower degree of dimensionality.

The \emph{w2v}-related approaches, which learn document embedding by averaging the word embedding, do not perform as well as our proposed approaches and the PVDM-related approaches that directly learn the document embedding. We also see that the overall performance on the novel dataset is better than that on the essay dataset, which is consistent with our expectation described in Section~\ref{sec:verification:dataset} and the observation in our previous work~\cite{ding2015visualizable}.

Considering the feature selection criteria for dynamic $n$-gram-based approaches, in this scenario the information gain measure outperforms the frequency measure, which is contradictory to the results reported in the survey~\cite{stamatatos2009survey}. The information-gain-based feature selection method consistently outperforms the frequency-based measure for the authorship verification problem.

\section{Evaluation on Authorship Characterization Problem}
\label{sec:characterization}
We evaluate the proposed models on another important problem in authorship analysis: authorship characterization. The problem is to identify the socio-linguistic characteristics of the author based on the given text. As discussed in Section~\ref{sec:introduction}, it has wide applications in marketing, political socialization, digital forensics, and social network analysis.

This problem can be described as follows: Given a set of labeled documents, where each document is assigned a class label such as the gender of its author, the problem is to identify the class label for a document whose author remains unknown. A classifier is trained on the labeled documents, and it assigns one of the labels to the targeted document. All labels are considered non-overlapping. For example, the labels for age ranges can be 18-23 and 23+.

To have the classifier understand the text data we need to represent these data in numeric form. To have the proposed models in this paper be able to solve this problem, we first learn the models based on the training text data by treating them as unlabeled data. After that we estimate the stylometric representations for all the documents in the training set and pad the available labels into the vectorized representations of the documents. Then an arbitrary classification model can be trained based on these vectors. Given the unseen document data, the models estimate their representations and feed them into the classifier to have the predicted labels.

In this section, we evaluate different stylometric representations for the authorship characterization problem. We represent the text documents in different forms based on the selected model and feed the learned representation of the documents into a simple logistic regression model to predict the characteristics of a text's author.

\begin{figure*}
\centering
\includegraphics[width=5in]{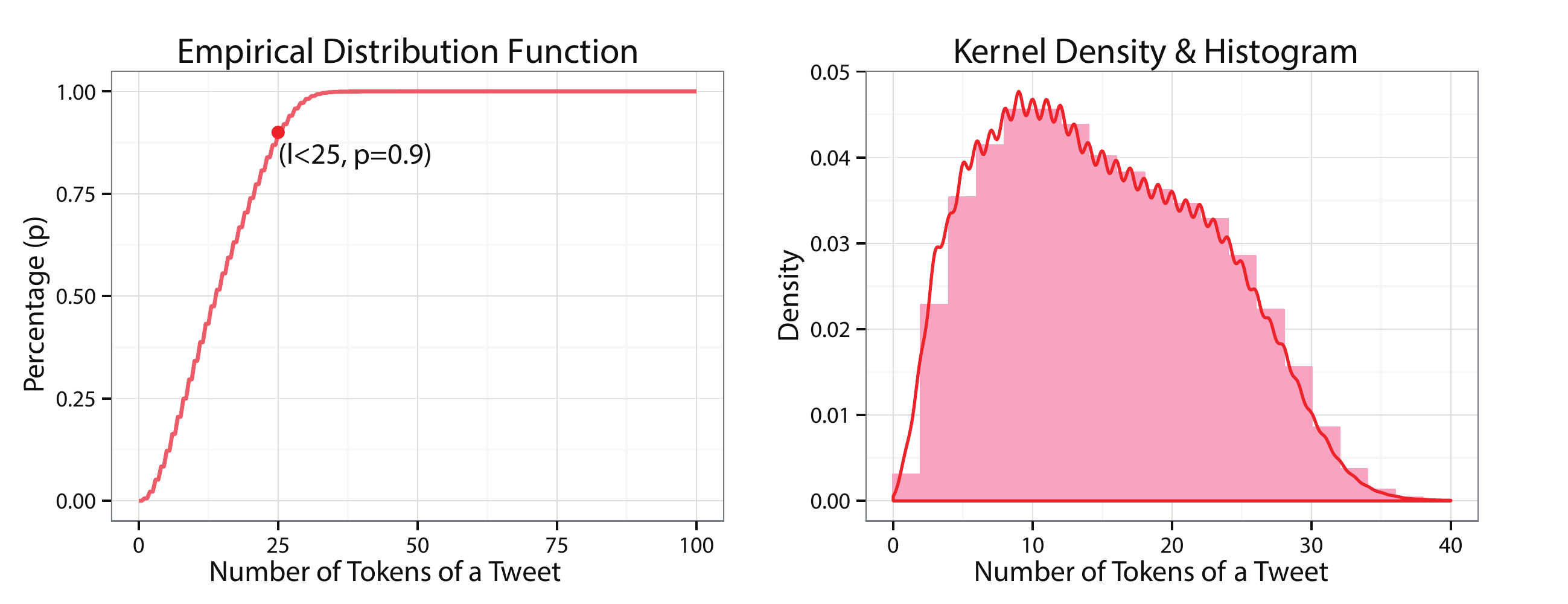}
 
\caption{Empirical distribution, kernel density and histogram on the tweets length for the ICWSM2012 Twitter dataset.}
\label{fig:characterization:dataset}
\end{figure*}

\newcolumntype{L}[1]{>{\raggedright\let\newline\\\arraybackslash\hspace{0pt}}m{#1}}
\begin{table}
\tbl{Summary of the cleaned ICWSM Twitter characterization dataset.\label{table:characterization:icwsm}}
{
\begin{tabular}{c|cccc}
\hline \hline
\textbf{Label type} & \textbf{Label} & \textbf{Users} & \textbf{Valid tweets} & \textbf{Tokens}\\ \hline
\multirow{2}{*}{\begin{tabular}[l]{@{}l@{}}Gender\end{tabular}}

& Female & 192 & 115,746 & 1,366,699  \\ \cline{3-5}
& Male & 192 & 127,368 & 1,475,018 \\ \hline
\multirow{2}{*}{\begin{tabular}[l]{@{}l@{}}Age\end{tabular}}

& (18 - 23) & 194 & 104,686 & 1,473,512  \\ \cline{3-5}
& (25 - 30) & 192 & 71,883 & 1,122,247 \\ \hline
\multirow{2}{*}{\begin{tabular}[l]{@{}l@{}}Political orientation\end{tabular}}

& Republican & 200 & 147,423 & 2,545,947  \\ \cline{3-5}
& Democrat & 200 & 170,822 & 2,957,180 \\ \hline  \hline
\end{tabular}
}
\end{table}

\subsection{The Twitter characterization dataset}

We choose the ICWSM 2012 labeled Twitter dataset~\cite{al2012homophily} in our experiment. This dataset consists of three categories of labels, and it is publicly available. Due to the limitation of Twitter's policy, the actual content of tweets were not included with the dataset; however, the Twitter users' identification numbers as well as their tweet IDs are available. We retrieve all the data using Twitter API according to the available information.

To preprocess the dataset, we remove all the non-ASCII characters and replace all the URLs with a special lexical token. We also pre-tokenize the tweets and assign POS tags for each token in each tweet using the pre-trained tagger from~\cite{ref:Twittertagger}. In this dataset there is other social-network-based information, such as the target user's friends, and the friends' tweets, etc. Since we only want to model the writing style of the Twitter user, we omit this information as well as those tweets that are re-tweeted by the given author. We attempt to include only the tweets that are authentically authored by the labeled Twitter user.

The labels in this dataset are generated semi-automatically and manually inspected~\cite{al2012homophily}. This dataset consists of three categories of labels for Twitter users: age, gender, and political orientation. The cleaned dataset is summarized in Table~\ref{table:characterization:icwsm}. There are 1170 Twitter users in total.

\begin{longitem}
\item \textit{Gender}. The labels for this category can be either \emph{male} or \emph{female}. The labels are automatically generated based on the Twitter user's name with a name-gender database, and then labels are manually inspected to ensure the labels are correct.

\item \textit{Age}.This dataset only distinguishes individuals of age ranges in \emph{18-23} or \emph{25-30}. It frames the age prediction into a binary classification problem. The labels are constructed by looking at the tweets about birthday announcement, e.g., ``Happy birthday to me".

\item \textit{Political orientation}. This dataset provides political Twitter users with a label: either \emph{Democrat} or \emph{Republican}. Twitter users are collected from the \emph{wefollow} Twitter directory~\cite{al2012homophily}.
\end{longitem}

Figure~\ref{fig:characterization:dataset} shows the empirical distribution, kernel density and histogram on the tweets' length in terms of lexical token size. In general, tweets are very short text snippets. 90 percent of tweets have less than 25 tokens and most have a length of around 10 tokens. We combine all tweets of a single user into a single document and treat each tweet as an individual sentence.

To proceed with the experiment we conduct a 10-fold cross validation on the Twitter dataset and collect the accuracy measure for each characterization approach. First we convert each document into its numeric vector representation using different stylometric representations in our proposed models or the baselines, and then we feed them into a simple logistic classifier to predict the label of the document.

\subsubsection{Baselines}
We inherit the same set of baselines as in the previous experiment on the authorship verification problem, except for those studies reported in PAN2014~\cite{rangel2014overview} since we do not have the available result for direct comparison. The baselines are configured to have the same hyper-parameter setting as the previous experiment. Additionally, we include several additional baselines:

\begin{longitem}
\item \emph{LDA}. In addition to picking the empirical optimal value $k=500$, which represents the latent topics, we include the performance of $ k \in \{100, 200, 500, 800\}$ for comparison.

\item \emph{LSA}. Likewise, for the LSA model we also include the performance of $k \in \{100, 200, 500, 800\}$ in addition to the original $k=200$. Recall that $k$ for LSA represents the number of singular values.

\item Moreover, we include two evaluation results that were presented by~\citeN{al2012homophily} since we follow the same setup and use the same dataset for the experiment. The \emph{target user info} approach is a SVM-based model trained on the features that are constructed on the user's tweets and other information. These features include textual features (e.g., stemmed $n$-grams and hash tags, etc.) and socio-linguistic features (e.g., re-tweeting tendency, neighborhood size, and mention frequency, etc.). The \emph{all info} approach is another SVM-based model trained on the features that adds additional social-network features (e.g., average of the neighborhood's feature vectors).
\end{longitem}

We notice that the baseline measures adopted from \cite{al2012homophily} have more advantages to our proposed approaches and our baseline approaches. First, they use a SVM model that typically outperforms a simple logistic regression model when the same data is given. Second, our approaches only consider the information reflected from the text, i.e., stylometric information. Other socio-linguistic, behavioral, and social-network-related information is discarded. Given the advantages of their approach we expect that probably they will outperform the others. However, our experiments show that our proposed model achieves even better accuracy, which will be described in the following section.

\begin{table*}
\centering
\tbl{Performance comparison for the authorship characterization problem on the ICWSM2012 Twitter dataset.\label{table:characterization:compare}}{
\begin{tabular}{l|rlllll}
\hline \hline
\begin{tabular}[l]{@{}l@{}}\textbf{Measure}\end{tabular}
 &
\begin{tabular}[l]{@{}l@{}}\textbf{Approach}\end{tabular}
 &
\begin{tabular}[l]{@{}l@{}}\textbf{Age}\end{tabular}
 &
\begin{tabular}[l]{@{}l@{}}\textbf{Gender}\end{tabular}
 &
\begin{tabular}[l]{@{}l@{}}\textbf{Political} \\ \textbf{Orientation}\end{tabular}
 &
\begin{tabular}[l]{@{}l@{}}\textbf{Average}\end{tabular}
 \\ \hline

\multirow{2}{*}{\begin{tabular}[l]{@{}l@{}}\\ \\ \\ Accuracy\end{tabular}}
&Modality [Lexical+Topical]	\textbf{*} & \textbf{0.7887}	& 0.8308	& \textbf{0.9318}	& \textbf{0.8504}\\ \cline{3-6}
&Modality [Topical]	\textbf{*} & 0.7606	& \textbf{0.8423}	& 0.9205	& 0.8411\\ \cline{3-6}
&Modality [Lexical]	\textbf{*} & 0.7782	& 0.8154	& 0.9148	& 0.8361\\ \cline{3-6}
&\cite{al2012homophily} (all info)	& 0.7720	& 0.8020	& 0.9150	& 0.8297\\ \cline{3-6}
&Modality [Character]	\textbf{*} & 0.7711	& 0.7846	& 0.9034	& 0.8197\\ \cline{3-6}
&\cite{al2012homophily} (target user only)	& 0.7510	& 0.7950	& 0.8900	& 0.8120\\ \cline{3-6}
&Static+[5000-freq-ngram]	& 0.7606	& 0.7615	& 0.8580	& 0.7934\\ \cline{3-6}
&Static+[2000-freq-ngram]	& 0.7500	& 0.7731	& 0.8523	& 0.7918\\ \cline{3-6}
&Static+[1500-freq-ngram]	& 0.7782	& 0.7308	& 0.8352	& 0.7814\\ \cline{3-6}
&PVDBOW+PVDM	& 0.7323	& 0.7346	& 0.8693	& 0.7787\\ \cline{3-6}
&w2v-skipgram	& 0.7112	& 0.7692	& 0.8380	& 0.7728\\ \cline{3-6}
&w2v-cbow+skipgram	& 0.7253	& 0.7692	& 0.8238	& 0.7728\\ \cline{3-6}
&w2v-cbow	& 0.7218	& 0.7807	& 0.8096	& 0.7707\\ \cline{3-6}
&Static+[1000-freq-ngram]	& 0.7465	& 0.7385	& 0.8097	& 0.7649\\ \cline{3-6}
&Static+[0500-freq-ngram]	& 0.7641	& 0.7192	& 0.8011	& 0.7615\\ \cline{3-6}
&LSA-k=200	& 0.6937	& 0.7577	& 0.8097	& 0.7537\\ \cline{3-6}
&LSA-k=100	& 0.6937	& 0.7538	& 0.8097	& 0.7524\\ \cline{3-6}
&LSA-k=500	& 0.7007	& 0.7500	& 0.8040	& 0.7516\\ \cline{3-6}
&PVDBOW	& 0.6936	& 0.6653	& 0.8409	& 0.7333\\ \cline{3-6}
&PVDM	& 0.6901	& 0.6653	& 0.8409	& 0.7321\\ \cline{3-6}
&Static+[0200-freq-ngram]	& 0.7570	& 0.7000	& 0.7301	& 0.7290\\ \cline{3-6}
&LDA-k=500	& 0.6338	& 0.7423	& 0.8040	& 0.7267\\ \cline{3-6}
&LDA-k=100	& 0.6303	& 0.7462	& 0.7869	& 0.7211\\ \cline{3-6}
&Static+[0100-freq-ngram]	& 0.7324	& 0.6846	& 0.7102	& 0.7091\\ \cline{3-6}
&Static+[1500-info-ngram]	& 0.7254	& 0.7000	& 0.6847	& 0.7034\\ \cline{3-6}
&Static+[5000-info-ngram]	& 0.6866	& 0.7231	& 0.6960	& 0.7019\\ \cline{3-6}
&Static+[2000-info-ngram]	& 0.7289	& 0.6962	& 0.6790	& 0.7014\\ \cline{3-6}
&Static	& 0.6904	& 0.7324	& 0.6769	& 0.6999\\ \cline{3-6}
&Static+[0500-info-ngram]	& 0.7394	& 0.6692	& 0.6847	& 0.6978\\ \cline{3-6}
&Static+[1000-info-ngram]	& 0.7113	& 0.7000	& 0.6818	& 0.6977\\ \cline{3-6}
&LDA-k=200	& 0.5986	& 0.7269	& 0.7585	& 0.6947\\ \cline{3-6}
&Static+[0200-info-ngram]	& 0.7183	& 0.6808	& 0.6818	& 0.6936\\ \cline{3-6}
&Static+[0100-info-ngram]	& 0.7289	& 0.6654	& 0.6761	& 0.6901 \\ \cline{3-6}
&Modality [syntactic]	\textbf{*} & 0.6303	& 0.6654	& 0.6364	& 0.6440

 \\ \hline \hline
\end{tabular}
}
\end{table*}

\subsubsection{Performance comparison}
The performance of our proposed models, as well as all the baselines, is listed in Table~\ref{table:characterization:compare}. It shows that our first proposed model, which jointly learns the representation for the lexical modality and the topical modality, achieves the highest accuracy value. The runner-up is the topical modality, and the character-level modality does not perform as well as the other two. The proposed lexical/topical modality model and the character-level modality model also outperform the PVDM-related models, \emph{w2v}-related models, and other dynamic $n$-gram-based models. Unlike the results for the authorship verification problem, the \emph{w2v}-related baselines perform fairly well. They achieve a higher accuracy value than PVDM, PVDBOW, LSA, and LDA.

Even the \emph{(target user only)} approach and the \emph{(all info)} approach are given advantage; however, it achieves a lower accuracy value than our proposed joint model for lexical and topical modality, which is contradictory to our expectation. It shows that our proposed approach better models the writing variation than the $n$-gram language model that is used in both of these two baselines.

Table~\ref{table:characterization:compare} also shows that the proposed syntactic representation learning model does not perform well on the ICWSM 2012 dataset, which is different from the previous authorship verification problem. This is because the tweet text data are relatively more casual than essay and novel, which does not introduce much variation in the grammatical bias.

Regarding the feature selection measure, in this scenario the frequency-based selection approach outperforms the information-gain-based selection-based approach. Even the top-100 frequency-ranked $n$-grams outperform top-$1500$ information-gain-ranked $n$-grams, which is completely different from the result shown in previous authorship verification experiments. Such a difference further confirms our argument that feature selection measures are scenario-dependent/data-dependent. Even the feature set is dynamically constructed based on a different dataset, the measurement for the selection process is data-dependent. A language model over text data is better than a feature-selection-based model.

\section{Conclusions and Future Directions}
\label{sec:conclusion}
In this article, we present our three models for learning the vectorized stylometric representations of different linguistic modalities for authorship analysis. To the best of our knowledge, it is the very first work introducing the problem of stylometric representation learning into the authorship analysis field. Our proposed models are designed to effectively capture the differences of writing styles of different modalities when an author is composing text. By using the proposed feature learning scheme, guided by the selected linguistic modality, we attempt to mitigate the issues related to the feature engineering process in current authorship study. Our experiments on the publicly available PAN 2014 and the ICWSM 2012 Twitter datasets, respectively, for the authorship verification problem and the authorship characterization problem, demonstrate that our proposed models are effective and robust on both different datasets and AA problems.

Our future research will focus on exploring better models to capture writing styles and proposing models for other languages. Currently the representation learning models are simple one-layer neural networks. A recurrent neural network with long-short term memory is more suitable for capturing the contextual relationship over long text. For learning the syntactic modality representation, a recursive neural network that operates on the fully parsed syntactic tree will fit more into the nature of grammatical variations than the current one. Moreover, this work only focuses on capturing the variations in English writing. Additional changes need to be applied for text in other languages.

\begin{acks}
 
The research is supported in part by the Discovery Grants (356065-2013) from the Natural Sciences and Engineering Research Council of Canada (NSERC), Canada Research Chairs Program (950-230623), and the Research Incentive Fund grant (RIF13059) from the Zayed University.\\
\end{acks}

\bibliographystyle{ACM-Reference-Format-Journals}
\bibliography{vea2}

\medskip

\end{document}